\documentclass{article}

\PassOptionsToPackage{numbers, sort}{natbib}

\usepackage[preprint]{neurips_2026}

\usepackage[utf8]{inputenc} % allow utf-8 input
\usepackage[T1]{fontenc}    % use 8-bit T1 fonts
\usepackage{hyperref}       % hyperlinks
\usepackage{url}            % simple URL typesetting
\usepackage{booktabs}       % professional-quality tables
\usepackage{amsfonts}       % blackboard math symbols
\usepackage{nicefrac}       % compact symbols for 1/2, etc.
\usepackage{microtype}      % microtypography
\usepackage{xcolor}         % colors

\usepackage{amsmath}
\usepackage{amssymb}
\usepackage{mathtools}
\usepackage{amsthm}
\usepackage{dsfont}
\usepackage{csquotes}
\usepackage{algorithm}
\usepackage{algorithmic}

\usepackage{caption}

\newtheorem{metric}{Metric}

\usepackage{tcolorbox}
\newtcolorbox{graybar}{
	boxrule=0pt,
	leftrule=4pt,
	colframe=gray!50,
	coltext=black!80, 
	colback=white,
	arc=0pt,
	left=6pt,
	right=6pt,
	top=0pt,
	bottom=0pt,
}
\definecolor{gold}{RGB}{205,150,50} % 金色

\usepackage{booktabs} % 专业的水平线
\usepackage{tabularx} % 自动调整列宽
\usepackage{multirow} % 跨行单元格
\usepackage{amsmath}  % 数学符号
\usepackage{xcolor}   % 颜色支持
\usepackage{nicematrix} % 更高级的表格线控制

\title{Reconciling Contradictory Views on the Effectiveness of SFT in LLMs: An Interaction Perspective}

\author{%
    Junpeng Zhang$^{1,2,3}$\,, Lei Cheng$^{1}$\,, Guoxi Zhang$^{2}$\,, Hua Cai$^{3}$\,, Qing Xu$^{3}$\,, Quanshi Zhang$^{1}$\thanks{Quanshi Zhang is the corresponding author. He is with the Department of Computer Science and Engineering, the John Hopcroft Center, at the Shanghai Jiao Tong University, China. \texttt{zqs1022@sjtu.edu.cn.}}
    \\
    $^1$Shanghai Jiao Tong University \ 
    $^2$Beijing Institute for General Artificial Intelligence \ $^3$UniDT \\
  {\small \texttt{\{zhangjp63, zqs1022\}@sjtu.edu.cn}}
}

\begin{document}

\maketitle

\begin{abstract}
This paper explores a scientific question in supervised fine-tuning (SFT): why SFT is broadly effective for small-scale deep neural networks, yet can produce inconsistent or even detrimental effects when applied to large language models (LLMs). Recent advances in interaction-based explanations~\citep{ren2024we} suggest that interactions between words/tokens provide a faithful metric for quantifying the inference patterns encoded by LLMs. We find that the evolution of interactions during SFT can effectively explain the inconsistent effectiveness of SFT for LLMs. Specifically, we find that (1) SFT primarily removes noise-like interactions, while rarely acquiring reliable new interactions. (2) This denoising stage is extremely brief, after which continued fine-tuning tends to introduce overfitted interactions. We validate these findings across multiple LLMs and datasets. Our findings provide new insights into early stopping and offer practical guidance for LLM training.
\end{abstract}

% note 
% 1. 在Figure 2：（1）caption中讲粗体e=[]^T加footnote；（2）在图中标注符号
% 2. 数值范围怎么讲
% 4. baseline设置
% 6. introduction三个（1）矛盾了，其次交互type和AND OR type是否矛盾了

\section{Introduction}
Supervised fine-tuning (SFT) has become a common strategy for adapting pretrained deep neural networks (DNNs) to domain-specific applications. However, its effectiveness in large language models (LLMs) has been subject to inconsistent and even contradictory viewpoints. While some prior studies~\citep{ouyang2022training, chung2024scaling, wei2021finetuned} considered SFT essential for improving instruction-following ability and overall usability, others~\citep{luo2025empirical, wang2022two, shi2024instruction, gudibande2023false} suggested that it may overfit to the fine-tuning data, limiting generalization and even degrading performance.

Therefore, rather than drawing a premature conclusion about the utility of SFT, this paper focuses on the following scientific question: \textbf{what internal factors account for the inconsistent effectiveness of SFT across different LLMs?} Specifically, we aim to quantify the inference patterns learned by LLMs during SFT, which help explain why SFT improves performance in some cases but induces overfitting in others.

\textbf{Background.} As an emerging explanation strategy, \citet{ren2024we} measure the interactions between input variables encoded by a DNN as primitive inference patterns. For example, the interactions in an LLM usually represent phrase patterns between different words/tokens.

\begin{figure*}[t]
    \centering
    \includegraphics[width=\textwidth]{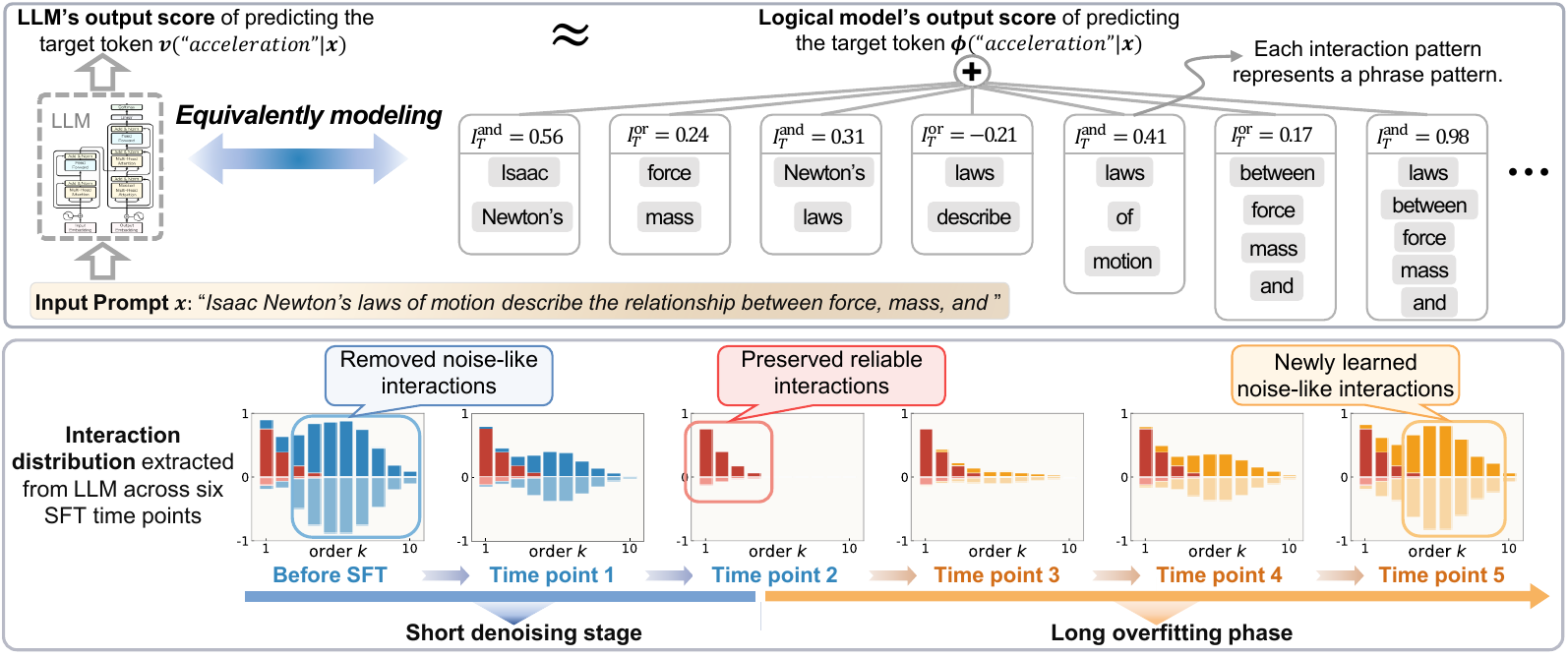}
    \vspace{-15pt}
    \caption{(Top) The complex inference patterns of an LLM can be faithfully represented by a few interactions (phrase patterns) in a logical model $\phi(\cdot)$. (Bottom) The SFT first enters a short denoising stage of removing noise-like interactions, which are non-generalizable and mutually canceling. In the subsequent prolonged overfitting stage, the LLM gradually learns new noise-like interactions again.}
    \vspace{-5pt}
    \label{img::fig1}
\end{figure*}

\noindent
\begin{graybar}As illustrated in Figure~\ref{img::fig1}, given a physics-related input prompt, an LLM may automatically learn a phrase pattern between the words \textit{``laws''}, \textit{``of''} and \textit{``motion,''} which can be quantified as an interaction. When all three words are present in the input prompt, this interaction is activated and contributes an effect of $0.41$ to boost the confidence in generating the token \textit{``acceleration.''} More importantly, a substantial body of empirical studies~\citep{li2023does, ren2023defining, ren2024we, chen2024defining, zhou2024explaining} and theoretical theorems\footnotemark have rigorously demonstrated the faithfulness of using interactions to explain the primitive inference patterns in LLMs.
\end{graybar}

\footnotetext{Please see Section~\ref{sec::preliminary} and Appendix~\ref{proof:match} for details.}

\textbf{Our work.} In this study, we find that the evolution of the representation quality of interactions can well explain the inconsistent effectiveness of SFT across different LLMs. Specifically, as illustrated in Figure~\ref{img::fig1}, we explore how new interactions emerge and how existing interactions are removed throughout the SFT process, yielding several key findings.

\textbf{(1) LLMs primarily remove noise interactions during SFT while rarely learning reliable ones.} 
We explore how the interactions encoded by an LLM evolve throughout the SFT process, and categorize the changed interactions into three types: \textit{removed interactions}, \textit{newly emerged interactions}, and \textit{preserved interactions}.

\textit{Removed interactions} refer to interactions that are eliminated during SFT.
We find that most removed interactions represent noise patterns, for three reasons. First, their positive and negative effects largely cancel out, \emph{i.e.,} approximately half of the removed interactions contribute positively to the prediction of target tokens, while the other half contribute negatively. Second, these interactions tend to exhibit higher complexity, where the complexity (order) of an interaction is quantified as the number of input variables involved in the interaction. Third, they show weak generalizability, meaning they are less likely to generalize across different LLMs.

\textit{Newly emerged interactions} are interactions that are newly acquired during SFT. Similar to removed interactions, most newly emerged interactions also correspond to non-generalizable noise patterns.

% Some interactions are newly acquired during SFT, which are termed the \textit{newly emerged interactions}. Similar to removed interactions, we find that most newly emerged interactions also represent non-generalizable noise patterns.

\textit{Preserved interactions} usually represent reliable patterns. They tend to have low complexity and contribute positively to the prediction of target tokens, with limited cancellation between positive and negative effects. Moreover, they exhibit relatively strong generalization across different LLMs.

\textbf{(2) The removal of noise interactions occurs only at the very early stage of SFT,} \textit{e.g.,} within the first $1000$ training steps, whereas further fine-tuning primarily introduces overfitted interactions. Moreover, we find that the emergence of these overfitted patterns is strongly correlated with an increasing gap between training and test losses in LLMs.

In summary, our findings suggest that SFT is indeed effective, yet its effective regime is surprisingly short. A small number of training samples is sufficient to remove a substantial fraction of noise interactions. Beyond this regime, continued fine-tuning is likely to introduce overfitted interactions. These results suggest that interactions can serve as diagnostic signals for monitoring SFT and provide a more principled criterion for early stopping in end-to-end SFT.

\section{Related Work}
\textbf{SFT of LLMs.} SFT has been widely used to adapt pretrained models to downstream tasks~\citep{devlin2019bert, raffel2020exploring}. 
Prior studies~\citep{ouyang2022training, wei2021finetuned, chung2024scaling, zhou2023lima} show that SFT can improve zero-shot generalization, instruction following, and task-specific performance. 
However, its effectiveness remains debated. SFT may over-specialize models to task formats and weaken their general in-context learning ability \citep{wang2022two}; it may imitate the response style of stronger models without acquiring their underlying capabilities \citep{gudibande2023false}; and continual instruction tuning may cause catastrophic forgetting of previously acquired knowledge and reasoning abilities \citep{luo2025empirical}. Furthermore, recent research has increasingly shifted attention toward alternative post-training paradigms, such as reinforcement learning from human feedback~\citep{dai2023safe, chai2024ma, ye2025robust}.

\textbf{Interaction-based explanation.}
Interaction-based explanation has recently attracted increasing attention for explaining DNNs, as surveyed in~\citep{zhou2025towards}. 
A large body of work studies how the complex inference logic of DNNs can be characterized by interactions~\citep{li2023does, ren2023defining, ren2024we, chen2024defining, zhou2024explaining}. 
For example, \citet{ren2024we} show that the output score of a DNN can be faithfully predicted by a small number of AND-OR interactions. 
\citet{zhou2024explaining} further use interactions to analyze the generalization power of DNNs, while \citet{wang2020unified} use interactions to explain adversarial transferability.

In this study, we show that interactions provide an interpretable and verifiable metric for characterizing SFT in LLMs. SFT first performs rapid denoising within a brief stage, but subsequent training is dominated by overfitting. This suggests that SFT for LLMs is necessary, yet challenges the expectation that large-scale fine-tuning on massive datasets consistently yields further gains.

\section{Understanding Representation Gain and Loss of LLM in SFT}
\subsection{Preliminaries: Interaction-Based Explanation}\label{sec::preliminary}
Given an LLM $v$ and an input prompt $\boldsymbol{x}$, which consists of $n$ input variables\footnote{\label{footnote::input-variables}In general, an input variable can be defined as the embedding vector(s) corresponding to a token, word, or phrase. In this paper, we follow~\citep{chen2024defining} to define each input variable as the embedding vector(s) of a word.} indexed by the set $N = \{1, 2, ..., n\}$, the LLM's output score $v(\boldsymbol{x}) \in \mathbb{R}$ is usually~\citep{chen2024evaluating, ren2024towards, deng2021discovering} defined as the following confidence score of generating a sequence of the target m tokens $[y_1, y_2,\cdots,y_m]$.

\noindent
{\small
\begin{align}\label{equation::vx}
    v(\boldsymbol{x}) \overset{\text{def}}{=} \sum_{i=1}^{m}\log \frac{p(y = y_i | \boldsymbol{x},\boldsymbol{y}_i^\text{preceding})}{1 - p(y = y_i | \boldsymbol{x},\boldsymbol{y}_i^\text{preceding})} \in \mathbb{R},
\end{align}}%
where {\small$\boldsymbol{y}_i^\text{preceding} \overset{\text{def}}{=} [y_1, y_2, \cdots, y_{i-1}]$} denotes the sequence of the preceding $(i-1)$ tokens before generating the $i$-th token. {\small$ p(y = y_i | \boldsymbol{x},\boldsymbol{y}_i^\text{preceding}) $} denotes the probability that the LLM predicts the target token $y_i$ at step $i$. Notably, {\small$\boldsymbol{y}_1^\text{preceding} = []$}.

Thus, the goal of interaction-based explanation is to disentangle a set of AND-OR interactions from the LLM as primitive inference patterns to compute the score $v(\boldsymbol{x})$. Specifically, \citet{chen2024defining} construct the following logical model composed of AND-OR interactions, where each interaction represents a phrase pattern automatically used by the LLM $v$ on a given input $\boldsymbol{x}$.

\noindent
{\small
\begin{equation}\label{equation::logical-model}
\begin{aligned}
   \!\! \forall \boldsymbol{x}' \in \Psi,\  \phi(\boldsymbol{x}')\overset{\text{def}}{=} 
\sum_{T \in \Omega^{\text{and}}} 
\underbrace{I^{\text{and}}_T \cdot \mathds{1}(\substack{\boldsymbol{x}'\  {\scriptstyle\text{triggers AND relation}} \\{\scriptstyle\text{between input variables in}} \ T})}_{\text{an AND interaction between input variables in }T} +\!\! \sum_{T \in \Omega^{\text{or}}} 
\underbrace{I^{\text{or}}_T \cdot \mathds{1}(\substack{\boldsymbol{x}'\  {\scriptstyle\text{triggers OR relation}} \\ {\scriptstyle\text{between input variables in}} \ T})}_{\text{an OR interaction between input variables in } T} + \ b,
\end{aligned}
\end{equation}}%
where $b$ is a scalar bias. $\Omega^{\text{and}}$ and $\Omega^{\text{or}}$ represent the set of AND interactions and the set of OR interactions, respectively. $I^{\text{and}}_T \in \mathbb{R}$ and $I^{\text{or}}_T \in \mathbb{R}$ denote numerical effects of interactions.

The AND trigger function {\small$ \mathds{1}(\substack{\boldsymbol{x}'\  {\scriptstyle\text{triggers AND relation}} \\{\scriptstyle\text{between input variables in}} \ T}) \in \{0, 1\} $} represents an AND relation between a set $ T \subseteq N$ of input variables\textsuperscript{\ref{footnote::input-variables}}. It returns $ 1 $ if all variables in $T$ are present (not masked\textsuperscript{\ref{footnote::masking}}) in $\boldsymbol{x}'$. The OR trigger function {\small$ \mathds{1}(\substack{\boldsymbol{x}'\  {\scriptstyle\text{triggers OR relation}} \\ {\scriptstyle\text{between input variables in}} \ T}) \in \{0, 1\} $} represents an OR relation between a set $ T \subseteq N $ of input variables\textsuperscript{\ref{footnote::input-variables}}. It returns $ 1 $ whenever any variable in $ T $ is present (not masked\textsuperscript{\ref{footnote::masking}}) in $\boldsymbol{x}'$. 

\textbf{Universal matching property.} Crucially, \citet{chen2024defining} propose a method to compute interaction effects $I^{\text{and}}_T$ and $I^{\text{or}}_T$ by analyzing the LLM output $v(\cdot)$ (see Appendix~\ref{sec:apdx-optimize-pq} for details and pseudocode). Furthermore, they prove that \textbf{the logical model $\phi$, constructed based on the extracted AND-OR interactions, can accurately predict the LLM outputs over all $2^n$ masked\footnote{\label{footnote::masking}An input word is masked by replacing the embedding vector(s) of its corresponding token(s) with a baseline vector $\boldsymbol{b}\in \mathbb{R}^d$. We follow~\citep{chen2024evaluating} to set the baseline vector $\boldsymbol{b}$ to represent a \textit{no-information} state of the input.}} states in $\Psi$.

\noindent
{\small
\begin{equation}\label{equation::universal-match}
\begin{aligned}
\forall \boldsymbol{x}' \in \Psi,\ \vert\phi(\boldsymbol{x}') - v(\boldsymbol{x}')\vert < \epsilon,
\end{aligned}
\end{equation}}%
where {\small$\Psi=\{\boldsymbol{x}_S | S\subseteq N\}$} denotes $2^n$ masked states of $\boldsymbol{x}$. $\boldsymbol{x}_S$ denotes a masked\textsuperscript{\ref{footnote::masking}} state that retains only the input variables in $S$, with all variables in $N\setminus S $ being masked\textsuperscript{\ref{footnote::masking}}.
$\epsilon$ is a small scalar to guarantee the fidelity of the explanations.

Please see Appendix~\ref{Appendix::examples-LLM} and Appendix~\ref{proof:match} for detailed empirical results on explaining LLMs and theoretical analyses demonstrating the high fidelity of interaction-based explanation, respectively. \textit{The supplementary material includes a \textbf{video demo} that showcases how interactions can be used to explain the inference patterns.}

\textbf{Sparsity of interactions.} \citet{ren2024we} prove that the number of interactions extracted by a DNN from a given input sample is relatively small and theoretically bounded, typically ranging from 50 to 150 interactions in practice. Please see Appendix~\ref{Appendix::interaction-sparsity} for empirical validation of the sparsity.

\textbf{Computational cost \& efficient solutions.} 
Similar to the computation of Shapley values~\citep{shapley1953value}, extracting interaction effects incurs exponential computational cost. However, prior studies have proposed a series of approximation algorithms~\citep{kang2024learning, kang2025spex, butler2025proxyspex}, as well as practical engineering techniques~\citep{li2023defining}, to enable efficient interaction extraction. These advances have made interaction-based explanations increasingly practical for debugging DNNs in various industrial applications~\citep{symtrustai2026}. Please see Appendix~\ref{Appendix::limitations} for details.

\subsection{Analyzing SFT of LLMs through Different Types of Interactions}
Conducting SFT on a target dataset has been regarded as an effective approach for adapting pre-trained DNNs~\citep{devlin2019bert, raffel2020exploring}. However, a growing debate has emerged regarding whether SFT remains consistently effective for LLMs. Some studies~\citep{luo2025empirical, wang2022two, shi2024instruction, gudibande2023false} argued that directly applying SFT to LLMs yields limited performance gains and may instead introduce unpredictable risks of overfitting.

% Therefore, in this study, \textbf{we aim to identify the internal factors that govern the improvement or degradation of representation quality in LLMs during SFT.} To this end, interaction has been widely considered~\citep{li2023does, ren2023defining, ren2024we, chen2024defining, zhou2024explaining} as a reliable metric for analyzing the factors (inference patterns) of a DNN.

Therefore, in this study, \textbf{we aim to identify the internal factors that account for the inconsistent effectiveness of SFT across different LLMs.} To this end, interaction has been widely considered~\citep{li2023does, ren2023defining, ren2024we, chen2024defining, zhou2024explaining} as a reliable metric for analyzing the internal factors (inference patterns) of a DNN. 

% Their reliability is supported by the following two facts.

% $\bullet$ It is proven that a DNN's prediction score on an input sample can be faithfully decomposed into a sum of numerical effects of a few interactions~\citep{chen2024defining}. Each interaction represents a primitive phrase pattern learned by the DNN.

% $\bullet$ It is found that simple interactions involving only a small number of tokens or words tend to represent reliable inference patterns that generalize well across models~\citep{zhou2024explaining}. In contrast, complex interactions are often associated with less reliable patterns that fail to generalize.

In this way, by continuously tracking interactions in LLMs throughout SFT, we uncover that (1) \textbf{the SFT primarily eliminates noise interactions while rarely acquiring reliable ones}, and (2) \textbf{this denoising stage is extremely brief, after which continued fine-tuning leads to overfitting}.

\subsubsection{Three Types: Newly emerged, Removed, and Preserved Interactions}
To better characterize the improvement and degradation of representation quality in LLMs during SFT, we categorize interactions into three types: (1) removed interactions, which are initially encoded by the LLM but subsequently eliminated by the SFT; (2) preserved interactions, which are retained throughout the SFT; and (3) newly emerged interactions, which are acquired during SFT.

Let $\Omega^{\text{and}}_{t}$ and $\Omega^{\text{or}}_{t}$ denote the sets of AND and OR interactions encoded by the LLM after $t$ gradient-update steps of SFT, respectively. In particular, $t=0$ denotes the time point before SFT. We then quantify the three types of interactions at each time point as follows.

$\bullet$ \textit{Type 1}: \textbf{Removed interactions} at time point $t$ refer to interactions that are encoded by the LLM before SFT but are eliminated within the first $t$ optimization steps\footnote{\label{footnote::optimization-step}An optimization step refers to one parameter-update operation during SFT, typically performed after computing gradients on one or more mini-batches of training samples.} of SFT. The set of removed interactions is identified as follows.

\noindent
{\small
\begin{equation}
\begin{aligned}
\forall \text{type} \in \{\text{and},\text{or}\},\quad \boldsymbol{R}^{\text{type}}_{t} \overset{\text{def}}{=} \Omega^{\text{type}}_0 \setminus  \Omega^{\text{type}}_t.
\end{aligned}
\end{equation}}%

$\bullet$ \textit{Type 2}: \textbf{Preserved interactions} at time point $t$ refer to interactions that are encoded by the LLM before SFT and remain encoded throughout the first $t$ optimization steps\textsuperscript{\ref{footnote::optimization-step}} of SFT. The set of preserved interactions is collected as follows.

\noindent
{\small
\begin{equation}
\begin{aligned}
\forall \text{type} \in \{\text{and},\text{or}\},\quad \boldsymbol{P}^{\text{type}}_{t} \overset{\text{def}}{=} \boldsymbol{P}^{\text{type}}_{t-1} \cap \Omega^{\text{type}}_t.
\end{aligned}
\end{equation}}%

$\bullet$ \textit{Type 3}: \textbf{Newly emerged interactions} at time point $t$ refer to interactions that are newly learned during the first $t$ optimization steps\textsuperscript{\ref{footnote::optimization-step}} of SFT. Newly emerged interactions are identified as the complement of the preserved interactions within all interactions encoded at time point $t$, as follows.

\noindent
{\small
\begin{equation}
\begin{aligned}
\forall \text{type} \in \{\text{and},\text{or}\},\quad \boldsymbol{E}^{\text{type}}_{t} \overset{\text{def}}{=} \Omega^{\text{type}}_t \setminus \boldsymbol{P}^{\text{type}}_{t}.
\end{aligned}
\end{equation}}%

To further assess the quality of newly emerged, removed, and preserved interactions during the SFT process, we introduce the following metrics to quantify the representational quality of interactions.

\textbf{Order/complexity of interactions.} The order (or complexity) of an interaction $T$ is defined as the number of input variables involved in $T$, \emph{i.e.}, $\text{order}(T) \overset{\text{def}}{=} \vert T \vert$. For example, as illustrated in Figure~\ref{img::fig1}, an interaction between two words (\emph{e.g.,} ``\textit{force}'' and ``\textit{mass}'') corresponds to a simple phrase pattern, whereas an interaction among five words (\emph{e.g.,} ``\textit{laws},'' ``\textit{between},'' ``\textit{force},'' ``\textit{mass}'' and ``\textit{and}'') represents a more complex phrase pattern.

It has been found that low-order interactions usually represent more reliable inference patterns than high-order interactions~\citep{ren2024towards, zhou2024explaining, ren2023bayesian}. (1) Low-order interactions are found to be more robust to input noise~\citep{ren2023bayesian}. (2) \citet{zhou2024explaining} demonstrate that low-order interactions exhibit stronger generalizability than high-order interactions.

\textbf{Generalizability of interactions.} 
Given a target LLM $v$ and an input prompt $\boldsymbol{x}$, we follow \citep{chen2024defining} to quantify the generalizability of the interactions encoded by $v$. An interaction is considered generalizable if it is consistently extracted across different DNNs. To this end, we introduce a baseline LLM $v'$ whose architecture differs from that of the target model. Specifically, if an interaction $S$ encoded by the target LLM $v$ is also captured by the baseline LLM $v'$, then this interaction $S$ is regarded as generalizable. Thus, the generalizability of an interaction $S$ can be quantified using the following binary metric.

\noindent
{\small
\begin{equation}
\begin{aligned}\label{equation::generalizability}
   \forall \text{type} \in \{\text{and},\ \text{or}\}, \forall S \in \Omega^{\text{type}}, \ \mathcal{G}^{\text{type}}_S \overset{\text{def}}{=} \mathds{1}(S \in \Omega^{\text{type}}_{v'})\cdot\mathds{1}(\text{sign}(I_S^{\text{type}}) = \text{sign}(I_{S, v^\prime}^{\text{type}})),
\end{aligned}
\end{equation}}%
where {\small$\mathds{1}(\cdot) \in \{0, 1\}$} is a trigger function that returns 1 if the given condition is satisfied. {\small$\Omega^{\text{and}}_{v'}$} and {\small$\Omega^{\text{or}}_{v'}$} denote the sets of AND and OR interactions encoded by the baseline LLM $v'$. {\small$I_{S, v^\prime}^{\text{and}}$} and {\small$I_{S, v^\prime}^{\text{or}}$} denote the scalar interaction effects encoded by the baseline LLM $v'$. {\small$\text{sign}(I_S^{\text{type}}) = \text{sign}(I_{S, v^\prime}^{\text{type}})$} indicates that interaction $S$ has a consistent effect on the prediction of the target token, meaning that it either increases or decreases the prediction score of the target token in both LLMs.

\textbf{Distribution of interactions.}
Instead of analyzing the complexity of each interaction individually, we characterize the distribution of interactions over different orders. For example, the distribution of newly emerged positive interactions over different orders and that of newly emerged negative interactions are represented by two vectors, {\small$\boldsymbol{e}^{+}=[\text{e}^{(1),+},\text{e}^{(2),+},\cdots,\text{e}^{(n),+}]^T$} and {\small$\boldsymbol{e}^{-}=[\text{e}^{(1),-},\text{e}^{(2),-},\cdots,\text{e}^{(n),-}]^T$}, respectively. $\text{e}^{(k),+}$ and $\text{e}^{(k),-}$ aggregate all emerged positive interactions of the $k$-th order and all emerged negative interactions of the $k$-th order, respectively.

\noindent
{\small
\begin{equation}\label{equation::all-strength}
\begin{aligned}
\text{e}^{(k),+}=f^{(k),+}_{\text{strength}} (\boldsymbol{E}^{\text{and}}_{t}, \boldsymbol{E}^{\text{or}}_{t}) &= \sum\nolimits_{\text{type} \in \{\text{and}, \text{or}\}} 
\sum\nolimits_{\substack{S \in \boldsymbol{E}^{\text{type}}_t:|S| = k}} \max(I^{\text{type}}_S, 0), 
\\ 
\text{e}^{(k),-}=f^{(k),-}_{\text{strength}} (\boldsymbol{E}^{\text{and}}_{t}, \boldsymbol{E}^{\text{or}}_{t}) &= \sum\nolimits_{\text{type} \in \{\text{and}, \text{or}\}} 
\sum\nolimits_{\substack{S \in \boldsymbol{E}^{\text{type}}_t:|S| = k}} \min(I^{\text{type}}_S, 0).
\end{aligned}
\end{equation}}%

Similarly, the distributions of removed interactions are denoted by 
{\small$\boldsymbol{r}^{+} = [\text{r}^{(1),+}, \text{r}^{(2),+}, \cdots, \text{r}^{(n),+}]^T$} 
and 
{\small$\boldsymbol{r}^{-} = [\text{r}^{(1),-}, \text{r}^{(2),-}, \cdots, \text{r}^{(n),-}]^T$}, respectively\footnote{\label{footnote-dirtribution-element}Elements in $\boldsymbol{r}^{+}$ and $\boldsymbol{r}^{-}$ are defined as 
{\scriptsize $\text{r}^{(k),+}=f^{(k),+}_{\text{strength}}(\boldsymbol{R}^{\text{and}}_{t}, \boldsymbol{R}^{\text{or}}_{t})$} 
and  
{\scriptsize $\text{r}^{(k),-}=f^{(k),-}_{\text{strength}}(\boldsymbol{R}^{\text{and}}_{t}, \boldsymbol{R}^{\text{or}}_{t})$}, 
respectively. 
Similarly, elements in $\boldsymbol{p}^{+}$ and $\boldsymbol{p}^{-}$ are defined as 
{\scriptsize $\text{p}^{(k),+}=f^{(k),+}_{\text{strength}}(\boldsymbol{P}^{\text{and}}_{t}, \boldsymbol{P}^{\text{or}}_{t})$} 
and 
{\scriptsize $\text{p}^{(k),-}=f^{(k),-}_{\text{strength}}(\boldsymbol{P}^{\text{and}}_{t}, \boldsymbol{P}^{\text{or}}_{t})$}, 
respectively.}.
The distributions of preserved interactions are represented by 
{\small$\boldsymbol{p}^{+} = [\text{p}^{(1),+}, \text{p}^{(2),+}, \cdots, \text{p}^{(n),+}]^T$} 
and 
{\small$\boldsymbol{p}^{-} = [\text{p}^{(1),-}, \text{p}^{(2),-}, \cdots, \text{p}^{(n),-}]^T$},
respectively\textsuperscript{\ref{footnote-dirtribution-element}}.

\begin{figure*}[t]
    \centering
    \includegraphics[width=\textwidth]{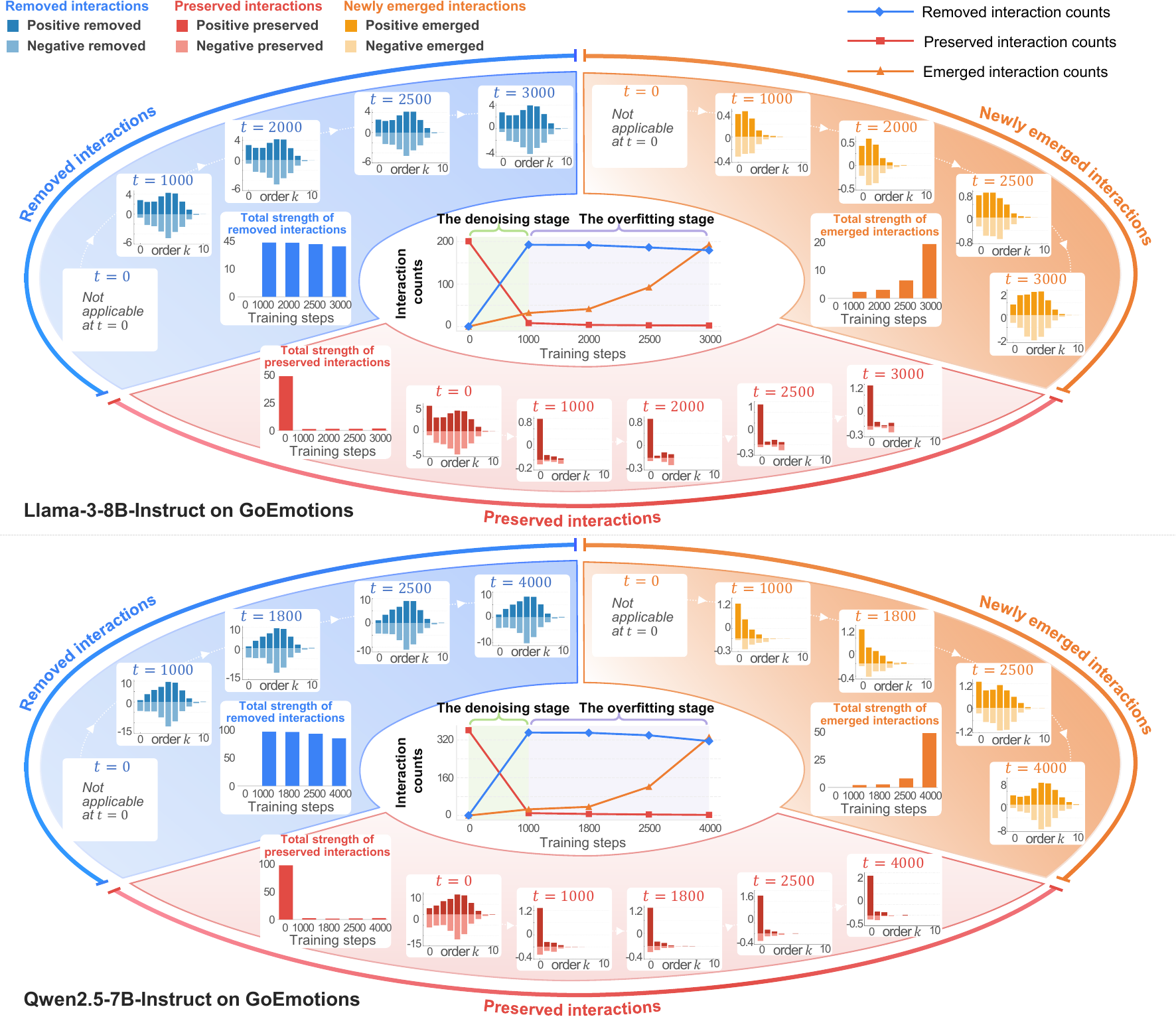}
    \vspace{-10pt}
    \caption{Evolution of the distribution of newly emerged interactions (\emph{i.e.,} {\small$\boldsymbol{e}^{+}$} and {\small$\boldsymbol{e}^{-}$}), removed interactions (\emph{i.e.,} {\small$\boldsymbol{r}^{+}$} and {\small$\boldsymbol{r}^{-}$}), and preserved interactions (\emph{i.e.,} {\small$\boldsymbol{p}^{+}$} and {\small$\boldsymbol{p}^{-}$}) throughout the SFT process. In the very short denoising stage of SFT, the LLM removes a large number of mutually canceling
    interactions, while only a small number of low-order interactions are preserved. In the later overfitting stage, many high-order mutually canceling interactions gradually emerge again. Additional experimental results on more LLMs are shown in Appendix~\ref{app:evolution-more-llms}.}
    \label{img::evolution-interactions}
\end{figure*}

\subsubsection{The Role of SFT: Denoising Rather than Reliable Representation Learning}
In this subsection, we conduct experiments to track the evolution of newly emerged, removed, and preserved interactions across the SFT process. We find that (1) LLMs primarily remove noise-like interactions during SFT, while rarely learning reliable interactions; (2) this denoising stage is extremely brief, after which continued fine-tuning leads to learning numerous overfitted patterns.

\textit{Experimental setting.}
We perform SFT on multiple LLMs across different datasets. For the GoEmotions dataset~\citep{demszky2020goemotions}, we fine-tune Qwen2.5-3B-Instruct~\citep{qwen2.5}, Qwen2.5-7B-Instruct~\citep{qwen2.5}, Llama-2-7B-Chat~\citep{touvron2023llama}, and Llama-3-8B-Instruct~\citep{grattafiori2024llama}. For the Unilaw-R1-Data dataset~\citep{cai2025unilaw}, we fine-tune Qwen2.5-3B-Instruct and Qwen2.5-7B-Instruct. For the Databricks-Dolly-15k dataset~\citep{DatabricksBlog2023DollyV2}, we fine-tune Gemma-3-4B-it~\citep{gemma_2025}. All fine-tuning experiments are conducted using LoRA, a parameter-efficient fine-tuning method~\citep{hu2022lora}. Please see Appendix~\ref{sec:training_setting} for detailed settings and dataset descriptions. Moreover, we follow~\citep{chen2024defining} to sample a set of words from the input prompt, where the embedding of each sampled word is treated as an input variable. Please see Appendix~\ref{sec:players} for details.

% with rank $r=8$, where LoRA adapters are applied to all eligible linear modules During SFT, the loss was computed only on the labeled response tokens, while the prompt tokens were excluded from the loss. We used a peak learning rate of $1\times10^{-4}$ with a warmup ratio of 0.1. We used a batch size of $1$ on each of the eight GPUs and accumulated gradients over 8 steps, resulting in an effective batch size of 64 samples per gradient update.

There are two commonly used metrics for evaluating the representation quality of interactions.
\begin{metric}[\textbf{Ratio of generalizable interactions}]
Given a set of AND-OR interactions $\Omega^{\text{and}}$ and $\Omega^{\text{or}}$, we quantify the proportion of generalizable interactions among all interactions, $\gamma(\Omega^{\text{and}}, \Omega^{\text{or}}) \in [0,100\%]$, as follows.
{\small
\begin{equation}
\begin{aligned}
\gamma(\Omega^{\text{and}}, \Omega^{\text{or}})
\overset{\text{def}}{=}
\frac{
\sum\nolimits_{\text{type} \in \{\text{and}, \text{or}\}}
\sum\nolimits_{\substack{S \in \Omega^{\text{type}}}}
|I^{\text{type}}_S \cdot \mathcal{G}^{\text{type}}_S|
}{
\sum\nolimits_{\text{type} \in \{\text{and}, \text{or}\}}
\sum\nolimits_{\substack{S \in \Omega^{\text{type}}}}
|I^{\text{type}}_S|
}
\times 100\%,
\end{aligned}
\end{equation}}%
where {\small$\mathcal{G}^{\text{type}}_S \in \{0,1\}$}, defined in Equation (\ref{equation::generalizability}), quantifies the generalizability of the interaction $S$.
\end{metric}

% denotes the boolean variable for the generalizability of an interaction.

\begin{metric}[\textbf{Ratio of uncancelled interaction effects}]
Given a set of AND-OR interactions $\Omega^{\text{and}}$ and $\Omega^{\text{or}}$, we quantify the ratio of interaction effects that remain uncancelled after the cancellation between positive and negative effects by
$\rho(\Omega^{\text{and}}, \Omega^{\text{or}}) \in [0,100\%]$, as follows.
{\small
\begin{equation}\label{equation::uncancelled}
\begin{aligned}
\rho(\Omega^{\text{and}}, \Omega^{\text{or}})
\overset{\text{def}}{=}
\frac{
\left|
\sum\nolimits_{\text{type} \in \{\text{and}, \text{or}\}}
\sum\nolimits_{S \in \Omega^{\text{type}}}
I^{\text{type}}_S
\right|
}{
\sum\nolimits_{\text{type} \in \{\text{and}, \text{or}\}}
\sum\nolimits_{S \in \Omega^{\text{type}}}
\left|I^{\text{type}}_S\right|
} \times 100\%.
\end{aligned}
\end{equation}}%
\end{metric}
If {\small$\rho(\Omega^{\text{and}}, \Omega^{\text{or}})$} is close to zero, it means that there is strong cancellation between positive and negative effects, suggesting that these interactions make little contribution to the prediction of target tokens and are more likely to reflect noise patterns.

\begin{figure*}[t]
    \centering
    \includegraphics[width=\textwidth]{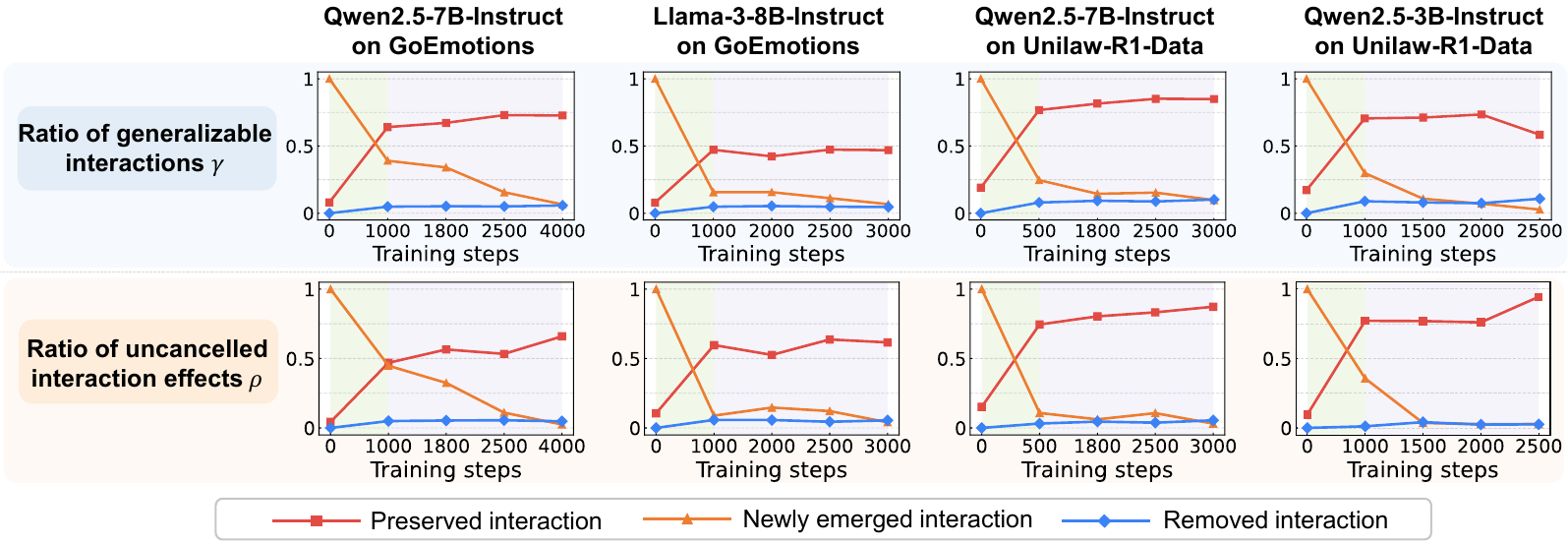}
    \vspace{-10pt}
    \caption{Evolution of the representation quality of newly emerged, removed, and preserved interactions during SFT, measured by generalizability $\gamma$ and uncancelled-effect ratio $\rho$. Newly emerged interactions show high quality only in the short denoising stage (the green region in the figure), but become less generalizable and more mutually canceling later (the purple region). Removed interactions remain low on both metrics throughout SFT, while preserved interactions improve after noise interactions are removed. Additional experimental results on more LLMs are shown in Appendix~\ref{app:quality-more-llms}.}
    \vspace{-5pt}
    \label{img::evolution-quality}
\end{figure*}

\textbf{(1) Analyzing newly emerged interactions.}
We track and analyze the representation quality of newly emerged interactions at different time points during SFT (\emph{e.g.,} the ratio of generalizable newly emerged interactions {\small$\gamma(\boldsymbol{E}^{\text{and}}_t, \boldsymbol{E}^{\text{or}}_t)$} at time point $t$). Figure~\ref{img::evolution-interactions} shows that the learning of newly emerged interactions during the SFT process can be divided into two stages.

\textbf{Finding 1:} \textit{LLMs learn only a few newly emerged interactions in the first stage, but begin to learn a large number of such interactions in the second stage.} Specifically, Figure~\ref{img::evolution-interactions} compares the interaction counts, {\small$\text{count}(\boldsymbol{E}^{\text{and}}_t, \boldsymbol{E}^{\text{or}}_t) = |\boldsymbol{E}^{\text{and}}_t| + |\boldsymbol{E}^{\text{or}}_t|$}, and the total interaction strength, {\small$\sum_{k=1}^{n}(|\text{e}^{(k), +}|+|\text{e}^{(k), -}|)$}, of interactions emerging in the first stage with those emerging in the second stage. The results show that only a small proportion of interactions emerge in the first denoising stage (green area in the figure), whereas the vast majority emerge in the second overfitting stage (purple area in the figure).

\textbf{Finding 2:} \textit{The small number of interactions that emerge in the first stage tend to be more generalizable, whereas the large number of interactions that emerge in the second stage behave more like noise patterns.} This claim is supported by three observations. (i) Figure~\ref{img::evolution-interactions} shows that interactions emerging in the second stage (\emph{i.e.,} the overfitting stage, corresponding to the purple region) are mostly high-order interactions, which are generally less reliable than those emerging in the first stage (\emph{i.e.,} the denoising stage, corresponding to the green region). (ii) Figure~\ref{img::evolution-quality} shows that interactions emerging in the first stage are more generalizable than those emerging in the second stage, indicating that most interactions emerging in the second stage are model-specific rather than shared across different DNNs. (iii) Figure~\ref{img::evolution-quality} further shows that interactions emerging in the first stage have a much higher ratio of uncancelled interaction effects than those emerging in the second stage.

These results suggest that LLMs tend to learn only a limited number of relatively reliable interactions in the first stage, while the huge number of newly emerged interactions in later stages are less likely to generalize and act like noise.

\textbf{(2) Analyzing removed interactions.} Figure~\ref{img::evolution-interactions} shows that the removal of interactions during the SFT process can also be divided into two stages.

\textbf{Finding 3:} \textit{The interaction removal in LLMs occurs primarily within an extremely short first stage.} The LLM removes the vast majority of interactions during the first stage, while only a negligible amount is removed in the second stage. Specifically, Figure~\ref{img::evolution-interactions} compares the interaction counts, {\small$\text{count}(\boldsymbol{R}^{\text{and}}_t, \boldsymbol{R}^{\text{or}}_t) = |\boldsymbol{R}^{\text{and}}_t| + |\boldsymbol{R}^{\text{or}}_t|$}, and the total interaction strength, {\small$\sum_{k=1}^{n}(|\text{r}^{(k), +}|+|\text{r}^{(k), -}|)$}, of interactions removed in the first stage with those removed in the second stage. The results show that most interaction removal occurs in the first stage, with only very few interactions being removed in the second stage.

\textbf{Finding 4:} \textit{The interactions removed during SFT predominantly correspond to noise patterns with poor generalizability.} This claim is supported by three observations. (i) Figure~\ref{img::evolution-interactions} shows that removed interactions are mostly high-order interactions, suggesting that they encode complex and less reliable patterns. (ii) Figure~\ref{img::evolution-quality} shows that almost none of the removed interactions are generalizable across different LLMs ({\small$\gamma(\boldsymbol{R}^{\text{and}}_t, \boldsymbol{R}^{\text{or}}_t) \approx 0$}), indicating that they are largely model-specific. (iii) Figure~\ref{img::evolution-quality} further shows that nearly all removed interaction effects exhibit mutual cancellation ({\small$\rho(\boldsymbol{R}^{\text{and}}_t, \boldsymbol{R}^{\text{or}}_t) \approx 0$}), suggesting that they contribute only marginally to target token prediction.

These results highlight one of the beneficial effects of SFT, \emph{i.e.,} it effectively removes unreliable/noise representations. However, this denoising process is short-lived, and continued fine-tuning beyond this stage does not further strengthen this effect.

\textbf{(3) Analyzing preserved interactions.} Figure~\ref{img::evolution-interactions} shows that the set of preserved interactions gradually stabilizes within the first $1000$ training steps. Afterward, the composition of these interactions remains largely unchanged, whereas their strengths continue to be reinforced during training.

\textbf{Finding 5:} \textit{LLMs retain only a small set of interactions during SFT, and these preserved interactions exhibit substantially higher representation quality than the other two types of interactions.} This claim is supported by three observations. (i) Figure~\ref{img::evolution-interactions} shows that preserved interactions are predominantly low-order interactions, suggesting that they represent simple and reliable inference patterns. (ii) Figure~\ref{img::evolution-quality} shows that, in most cases, more than half of the preserved interactions are generalizable across different LLMs, \emph{i.e.,} {\small$\gamma(\boldsymbol{P}^{\text{and}}_t, \boldsymbol{P}^{\text{or}}_t) > 50\%$}, indicating that they capture shared representations across LLMs. (iii) Figure~\ref{img::evolution-quality} further shows that preserved interactions exhibit weaker positive-negative cancellation, \emph{i.e.,} larger $\rho$ values, than the other two types of interactions, suggesting that they contribute more consistently to target token prediction.

These results suggest that one of the beneficial effects of SFT is to identify and reinforce high-quality interactions that already exist in the LLM.

\textbf{The primary utility of SFT lies in rapid denoising, followed by prolonged overfitting.} In summary, all these results suggest that during SFT, LLMs primarily remove noise-like interactions without learning many reliable interactions. Moreover, this denoising stage is extremely short. Once this stage ends, continued fine-tuning mainly introduces numerous noise-like interactions.

\subsection{Are Preserved Interactions the Backbone of LLM Inference?}
Since our previous findings suggest that the primary utility of SFT lies in denoising rather than learning new interactions, in this subsection, we further examine the following hypothesis: \textbf{the small set of interactions preserved in SFT constitutes the key backbone supporting target token prediction in LLMs}. In other words, we ask whether most interaction effects used for token prediction originate from this small set of preserved interactions, while the large number of removed interactions and interactions newly emerged in later stages contribute only marginally to the model’s predictions. We validate this hypothesis from the following three perspectives.

\textbf{(1) Verifying that removed and newly emerged interactions are predominantly noise, which is characterized by positive-negative cancellation of interaction effects.} We measure the ratio of uncancelled interaction effects $\rho$ (see Equation (\ref{equation::uncancelled})) of different types of interactions, including preserved, newly emerged, and removed interactions. 

Figure~\ref{img::evolution-quality} shows that preserved interactions exhibit the highest ratio of uncancelled effects, whereas newly emerged and removed interactions are largely characterized by positive-negative cancellation. This suggests that the latter predominantly capture noise patterns, while preserved interactions encode meaningful signals that contribute to prediction.

\textbf{(2) Verifying that preserved interactions exhibit a larger average contribution of individual interactions to target token prediction.} 
We measure the average contribution of individual removed interactions\footnote{\label{footnote::analyzing-three-types-interactions}Preserved and removed interactions are identified from the final LLM after the entire SFT. For newly emerged interactions, we separately evaluate interactions emerging during the denoising stage and those emerging in the subsequent stage.} as 
{\small$\sum_{k=1}^{n}(\text{r}^{(k), +}+\text{r}^{(k), -}) / (|\boldsymbol{R}^{\text{and}}_t| + |\boldsymbol{R}^{\text{or}}_t|)$}. 
The corresponding quantities for newly emerged\textsuperscript{\ref{footnote::analyzing-three-types-interactions}} and preserved interactions are computed as 
{\small$\sum_{k=1}^{n}(\text{e}^{(k), +}+\text{e}^{(k), -}) / (|\boldsymbol{E}^{\text{and}}_t| + |\boldsymbol{E}^{\text{or}}_t|)$} 
and 
{\small$\sum_{k=1}^{n}(\text{p}^{(k), +}+\text{p}^{(k), -}) / (|\boldsymbol{P}^{\text{and}}_t| + |\boldsymbol{P}^{\text{or}}_t|)$}, respectively.

\begin{figure*}[t]
    \centering
    \includegraphics[width=\textwidth]{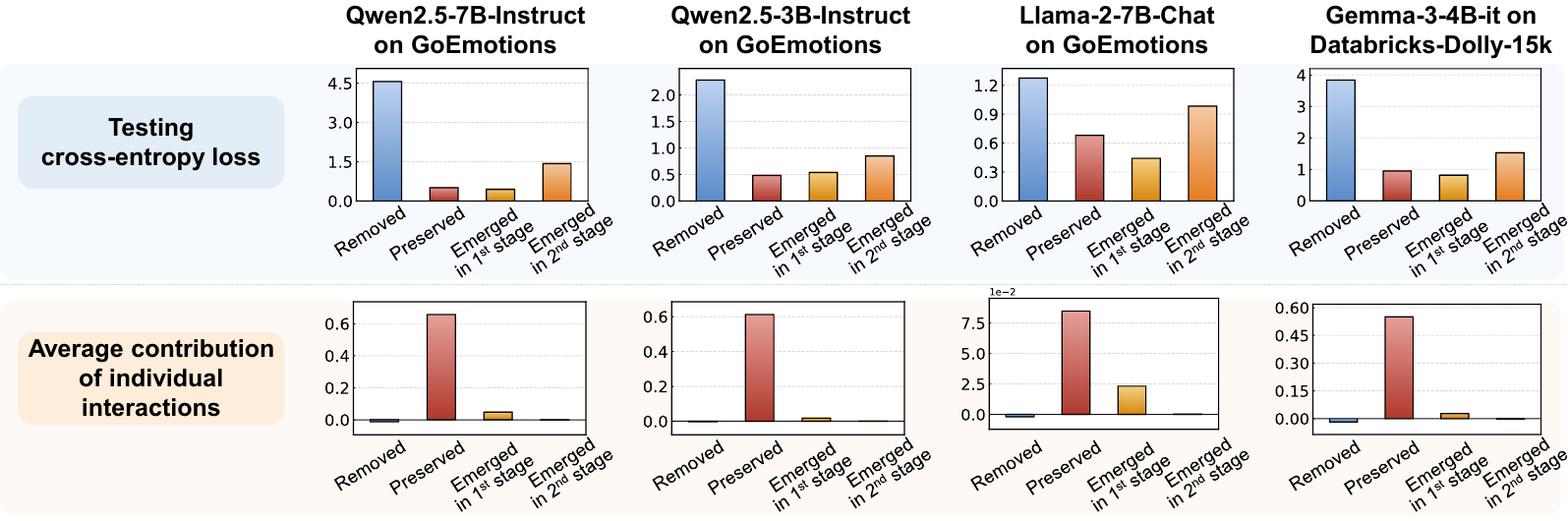}
    \vspace{-10pt}
    \caption{Prediction utility and individual contributions of different types of interactions. We compare removed interactions, preserved interactions, early-emerged interactions in the denoising stage, and later-emerged interactions in the overfitting stage. Preserved and early-emerged interactions yield lower test loss and larger per-interaction contributions, whereas removed and later-emerged interactions contribute only marginally. These results suggest that preserved interactions, together with a few early-emerged interactions, form the main backbone of LLM inference.}
    \vspace{-5pt}
    \label{img::backbone}
\end{figure*}

Figure~\ref{img::backbone} shows that preserved interactions and early-emerged interactions in the denoising stage contribute substantially more to target token prediction than removed interactions and later-emerged interactions in the overfitting stage. In contrast, the weak contributions of removed and later-emerged interactions suggest that they are more likely to represent noise patterns.

\textbf{(3) Verifying that preserved interactions are sufficient for basic inference.}
We evaluate the contribution of different types of interactions to target token prediction. In the setting of single-token generation, the prediction probability of the target token $y^*$ can be recovered from the output score as 
$p(y^{*}|\boldsymbol{x}) = g^{-1}(v(\boldsymbol{x}))$\footnote{The model output score on a given input prompt $\boldsymbol{x}$ can be written as {\scriptsize$v(\boldsymbol{x}) = g(p(y^{*}|\boldsymbol{x})) = \log( p(y^{*}|\boldsymbol{x})/(1-p(y^{*}|\boldsymbol{x})))$}.}, according to Equation (\ref{equation::vx}). Because the output score of an LLM can be decomposed into the sum of all interaction effects (see Equation (\ref{equation::universal-match})), the following probability naturally measures the utility of using only preserved interactions for token prediction.
\noindent
{\small
\begin{equation}
\begin{aligned}
p_{\text{preserved}}(y^*|\boldsymbol{x}) = g^{-1}(\sum_{k=1}^n \text{p}^{(k), +} + \text{p}^{(k), -}).
\end{aligned}
\end{equation}}%
Similarly, {\small$p_{\text{removed}}(y^*|\boldsymbol{x}) = g^{-1}(\sum_{k=1}^n \text{r}^{(k), +} + \text{r}^{(k), -})$} and {\small$p_{\text{emerged}}(y^*|\boldsymbol{x}) = g^{-1}(\sum_{k=1}^n \text{e}^{(k), +} + \text{e}^{(k), -})$} measure the utilities of using only removed interactions and newly emerged interactions for token prediction, respectively.

Figure~\ref{img::backbone} compares the cross-entropy loss on test samples when target token prediction is performed using only preserved interactions, removed interactions, early-emerged interactions\textsuperscript{\ref{footnote::analyzing-three-types-interactions}} in the denoising stage, or later-emerged interactions in the overfitting stage. Specifically, the losses are computed from the prediction probabilities induced by each type of interaction\textsuperscript{\ref{footnote::analyzing-three-types-interactions}}. The results show that using only preserved interactions or early-emerged interactions yields lower loss than using only removed interactions or later-emerged interactions. This indicates that preserved interactions and a few early-emerged interactions form the backbone supporting target token prediction.

Taken together, the above three sets of experiments provide consistent evidence that preserved interactions form the core backbone underlying target token prediction in LLMs.

\section{Conclusion and Discussion}
In this paper, we use interactions to address the scientific question of the inconsistent effectiveness of SFT on LLMs. We find that: (1) SFT primarily removes noise-like interactions from LLMs, while rarely inducing reliable new interactions; and (2) this denoising stage is extremely brief, after which continued fine-tuning leads to the emergence of numerous overfitted patterns. We validate these findings across multiple LLMs and datasets. These results help explain why SFT can sometimes exhibit inconsistent or even detrimental effects when applied to LLMs.

Our findings suggest that interactions can serve as diagnostic signals for monitoring SFT and provide a more principled criterion for early stopping in end-to-end SFT. In practice, our study challenges the conventional belief that fine-tuning LLMs on massive datasets is necessarily beneficial. The extremely brief denoising stage suggests that collecting large-scale SFT data may offer limited additional value. Moreover, our method provides a way to identify the early stopping point for a given LLM.

\newpage

\bibliography{neurips}
\bibliographystyle{plainnat}

%%%%%%%%%%%%%%%%%%%%%%%%%%%%%%%%%%%%%%%%%%%%%%%%%%%%%%%%%%%%
\newpage

\appendix
\section*{Appendix}
This appendix provides detailed information that supports the main paper. For clarity, the appendix is organized as follows.

\begin{itemize}
  \item Appendix~\ref{proof:match} provides validation of the high fidelity of interaction-based explanations.
  \item Appendix~\ref{Appendix::interaction-sparsity} provides validation of the sparsity of interactions.
  \item Appendix~\ref{Appendix::limitations} discusses the limitations of this work.
  \item Appendix~\ref{Appendix-more-results} reports additional experimental results.
  \item Appendix~\ref{sec:experimental_setting} presents experimental details.
  \item Appendix~\ref{sec:apdx-optimize-pq} describes the procedure for extracting the sparsest AND-OR interactions.
  \item Appendix~\ref{sec:impacts} discusses the broader impacts of this work.
\end{itemize}

% \begin{itemize}
%   \item Appendix~\ref{proof:match} validates the high fidelity of the interaction-based explanations.
%   \item Appendix~\ref{appendix::loss} describes the implementation details for the transferability-boosted loss $\mathcal{L}_{\text{trans}}$.
%   \item Appendix~\ref{Appendix::interaction-sparsity} examines the sparsity properties of interactions.
%   \item Appendix~\ref{appendix::practical-value} discusses the practical value of this work.
%   \item Appendix~\ref{Appendix-more-results} presents more experimental results that complement those reported in the main paper.
%   \item Appendix~\ref{sec:experimental_setting} provides detailed experimental settings.
%   \item Appendix~\ref{sec:apdx-optimize-pq} introduces the procedure for extracting interactions from DNNs.
% \end{itemize}

\section{Validation of High-Fidelity Interaction-Based Explanation}
\label{proof:match}
In this section, we validate the high fidelity of the interaction-based explanation from both theoretical and empirical perspectives. Specifically, \citet{chen2024defining} have theoretically proven that the logical model $\phi$, constructed from the extracted AND-OR interactions, can accurately predict the LLM outputs for all $2^n$ masked states of $\boldsymbol{x}$. Therefore, for completeness, we first review this existing theoretical guarantee to establish the fidelity of $\phi$, and then further corroborate it through our empirical validation.

\subsection{Theoretical Validation}
\begin{proof} \textbf{(1) Universal matching theorem of AND interactions.}

We will prove that output component \( u^{\text{and}}_S \) on all \( 2^n \) masked samples \( \{\boldsymbol{x}_S:S\subseteq N\} \) could be universally explained by all interactions in \( S\subseteq N \), \emph{i.e.}, \( \forall \emptyset \neq S\subseteq N, u^{\text{and}}_S = \sum_{\emptyset \neq T\subseteq S} I^{\text{and}}_T + v(\boldsymbol{x}_\emptyset) \). In particular, we define \( v^{\text{and}}_\emptyset = v(\boldsymbol{x}_\emptyset) \) (\textit{i.e.}, we attribute output on an empty sample to AND interactions).

Specifically, the AND interaction is defined as \( I^{\text{and}}_T = \sum\nolimits_{L \subseteq T} (-1)^{|T|-|L|} u^{\text{and}}_L \). 
To compute the sum of AND interactions \( \sum_{\emptyset \neq T\subseteq S} I^{\text{and}}_T = \sum\nolimits_{\emptyset \neq T \subseteq S} \sum\nolimits_{L \subseteq T} (-1)^{\vert T \vert - \vert L \vert} u^{\text{and}}_L \), we first exchange the order of summation of the set \( L\subseteq T\subseteq S \) and the set \( T \supseteq L \). 
That is, we compute all linear combinations of all sets \( T \) containing \( L \) with respect to the model outputs \( u^{\text{and}}_L \) given a set of input variables \( L \), \textit{i.e.}, \( \sum\nolimits_{T: L \subseteq T \subseteq S} (-1)^{|T|-|L|} u^{\text{and}}_L \). 
Then, we compute all summations over the set \( L\subseteq S \).

In this way, we can compute them separately for different cases of \( L\subseteq T\subseteq S \). In the following, we consider the cases (1) \( L = S = T \), and (2) \( L\subseteq T\subseteq S, L\ne S \), respectively.

(1) When \( L=S=T \), the linear combination of all subsets \( T \) containing \( L \) with respect to the model output \( u^{\text{and}}_L \) is \( (-1)^{|S|-|S|} u^{\text{and}}_L = u^{\text{and}}_L \).

(2) When \( L\subseteq T\subseteq S, L\ne S \), the linear combination of all subsets \( T \) containing \( L \) with respect to the model output \( u^{\text{and}}_L \) is \( \sum\nolimits_{T: L \subseteq T \subseteq S} (-1)^{|T|-|L|} u^{\text{and}}_L \). For all sets \( T: S\supseteq T\supseteq L \), let us consider the linear combinations of all sets \( T \) with number \( |T| \) for the model output \( u^{\text{and}}_L \), respectively. Let \( m := |T| - |L| \), (\( 0\le m\le |S|-|L| \)), then there are a total of \( C_{|S|-|L|}^{m} \) combinations of all sets \( T \) of order \( |T| \). Thus, given \( L \), accumulating the model outputs \( u^{\text{and}}_L \) corresponding to all \( T\supseteq L \), then \( \sum\nolimits_{T: L \subseteq T \subseteq S} (-1)^{|T|-|L|} u^{\text{and}}_L = u^{\text{and}}_L \cdot \underbrace{\sum\nolimits_{m=0}^{\vert S \vert - \vert L \vert} C_{|S|-|L|}^m (-1)^m}_{=0} = 0 \). Please see the complete derivation of the following formula.

\begin{equation}\begin{aligned}
    \sum\nolimits_{\emptyset \neq T \subseteq S} I^{\text{and}}_T
    = &  \sum\nolimits_{\emptyset \neq T \subseteq S} \sum\nolimits_{L \subseteq T} (-1)^{\vert T \vert - \vert L \vert} u^{\text{and}}_L \\
    = & \sum\nolimits_{L \subseteq S} \sum\nolimits_{T: L \subseteq T \subseteq S} (-1)^{\vert T \vert - \vert L \vert} u^{\text{and}}_L  - v^{\text{and}}_\emptyset \\
    = & \underbrace{u^{\text{and}}_S}_{L = S} + \sum\nolimits_{L \subseteq S, L \neq S} u^{\text{and}}_L \cdot \underbrace{\sum\nolimits_{m=0}^{\vert S \vert - \vert L \vert} C_{|S|-|L|}^m (-1)^m}_{=0}   - v^{\text{and}}_\emptyset \\
     = & u^{\text{and}}_S  - v^{\text{and}}_\emptyset  = u^{\text{and}}_S  - v(\boldsymbol{x}_\emptyset)
\end{aligned}\end{equation}

Thus, we have \( \forall \emptyset \neq S\subseteq N, u^{\text{and}}_S = \sum_{\emptyset \neq T\subseteq S} I^{\text{and}}_T + v(\boldsymbol{x}_\emptyset) \).

\textbf{(2) Universal matching theorem of OR interactions.}

According to the definition of OR interactions, we will derive that \( \forall S\subseteq N, u^{\text{or}}_S = \sum_{T:T\cap S\neq \emptyset} I^{\text{or}}_T \), 
where we define \( v^{\text{or}}_\emptyset = 0 \) (recall that in Step (1), we attribute the output on empty input to AND interactions).

Specifically, the OR interaction is defined as \( I^{\text{or}}_T = -\sum\nolimits_{L \subseteq T} (-1)^{|T|-|L|} v^{\text{or}}_{N\setminus L} \).
Similar to the above derivation of the universal matching theorem of AND interactions, to compute the sum of OR interactions \( \sum\nolimits_{T:T \cap S \neq \emptyset} I^{\text{or}}_T = \sum\nolimits_{T:T \cap S \neq \emptyset} \left[- \sum\nolimits_{L \subseteq T} (-1)^{\vert T \vert - \vert L \vert} v^{\text{or}}_{N \setminus L} \right] \), we first exchange the order of summation of the set \( L\subseteq T \subseteq N \) and the set \( T:T \cap S \neq \emptyset \). That is, we compute all linear combinations of all sets \( T \) containing \( L \) with respect to the model outputs \( v^{\text{or}}_{N \setminus L} \) given a set of input variables \( L \), \textit{i.e.}, \( \sum\nolimits_{T: T \cap S \neq \emptyset, T \supseteq L} (-1)^{\vert T \vert - \vert L \vert} v^{\text{or}}_{N \setminus L} \). Then, we compute all summations over the set \( L\subseteq N \).

In this way, we can compute them separately for different cases of \( L\subseteq T\subseteq N, T \cap S \neq \emptyset \). In the following, we consider the cases (1) \( L = N \setminus S \), (2) \( L=N \), (3) \( L \cap S \neq \emptyset, L \neq N \), and (4) \( L \cap S=\emptyset, L \neq N \setminus S \), respectively.

(1) When \( L = N \setminus S \), the linear combination of all subsets \( T \) containing \( L \) with respect to the model output \( v^{\text{or}}_{N \setminus L} \) is \( \sum\nolimits_{T: T \cap S \neq \emptyset, T \supseteq L} (-1)^{\vert T \vert - \vert L \vert} v^{\text{or}}_{N \setminus L} = \sum\nolimits_{T: T \cap S \neq \emptyset, T \supseteq L} (-1)^{\vert T \vert - \vert L \vert} u^{\text{or}}_S \). For all sets \( T: T\supseteq L, T \cap S \neq \emptyset \) (then \( T \neq N \setminus S, T \neq L \)), let us consider the linear combinations of all sets \( T \) with number \( |T| \) for the model output \( u^{\text{or}}_S \), respectively. Let \( |T'| := |T| - |L| \), (\( 1\le |T'|\le |S| \)), then there are a total of \( C_{|S|}^{|T'|} \) combinations of all sets \( T' \) of order \( |T'| \). 
Thus, given \( L \), accumulating the model outputs \( u^{\text{or}}_S \) corresponding to all \( T\supseteq L \), then \( \sum\nolimits_{T: T \cap S \neq \emptyset, T \supseteq L} (-1)^{\vert T \vert - \vert L \vert} v^{\text{or}}_{N \setminus L} = u^{\text{or}}_S \cdot \underbrace{\sum\nolimits_{|T'|=1}^{\vert S \vert } C_{|S|}^{|T'|} (-1)^{|T'|}}_{=-1} = -u^{\text{or}}_S \).

(2) When \( L=N \) (then \( T=N \)), the linear combination of all subsets \( T \) containing \( L \) with respect to the model output \( v^{\text{or}}_{N \setminus L} \) is \( \sum\nolimits_{T: T \cap S \neq \emptyset, T \supseteq L} (-1)^{\vert T \vert - \vert L \vert} v^{\text{or}}_{N \setminus L} = (-1)^{\vert N \vert - \vert N \vert} v^{\text{or}}_\emptyset = v^{\text{or}}_\emptyset \).

(3) When \( L \cap S \neq \emptyset, L \neq N \), the linear combination of all subsets \( T \) containing \( L \) with respect to the model output \( v^{\text{or}}_{N \setminus L} \) is \( \sum\nolimits_{T: T \cap S \neq \emptyset, T \supseteq L} (-1)^{\vert T \vert - \vert L \vert} v^{\text{or}}_{N \setminus L} \). For all sets \( T: T\supseteq L, T \cap S \neq \emptyset \), let us consider the linear combinations of all sets \( T \) with number \( |T| \) for the model output \( u^{\text{or}}_S \), respectively. Let us split \( |T| - |L| \) into \( |T'| \) and \( |T''| \), \textit{i.e.}, \( |T| - |L| = |T'| + |T''| \), where \( T'=\{i|i\in T, i\notin L, i\in N\setminus S\} \), \( T''=\{i|i\in T, i\notin L, i\in S\} \) (then \( 0\le|T''|\le|S|-|S\cap L| \)) and \( |T'| + |T''| + |L| = |T| \). In this way, there are a total of \( C_{|S|-|S\cap L|}^{|T''|} \) combinations of all sets \( T'' \) of order \( |T''| \). Thus, given \( L \), accumulating the model outputs \( v^{\text{or}}_{N\setminus L} \) corresponding to all \( T\supseteq L \), then \( \sum\nolimits_{T: T \cap S \neq \emptyset, T \supseteq L} (-1)^{\vert T \vert - \vert L \vert} v^{\text{or}}_{N \setminus L} \!=\! v^{\text{or}}_{N \setminus L} \cdot \sum_{T' \subseteq N\setminus S \setminus L} \underbrace{\sum\nolimits_{\vert T'' \vert = 0}^{\vert S \vert-\vert S \cap L \vert} C_{\vert S \vert - \vert S \cap L \vert}^{\vert T''\vert } (-1)^{\vert T' \vert + \vert T'' \vert} }_{=0} \!=\! 0 \).

(4) When \( L \cap S=\emptyset, L \neq N \setminus S \), the linear combination of all subsets \( T \) containing \( L \) with respect to the model output \( v^{\text{or}}_{N \setminus L} \) is \( \sum\nolimits_{T: T \cap S \neq \emptyset, T \supseteq L} (-1)^{\vert T \vert - \vert L \vert} v^{\text{or}}_{N \setminus L} \). Similarly, let us split \( |T| - |L| \) into \( |T'| \) and \( |T''| \), \textit{i.e.}, \( |T| - |L| = |T'| + |T''| \), where \( T'=\{i|i\in T, i\notin L, i\in N\setminus S\} \), \( T''=\{i|i\in T, i\in S\} \) (then \( 0\le|T''|\le|S| \)) and \( |T'| + |T''| + |L| = |T| \). In this way, there are a total of \( C_{|S|}^{|T''|} \) combinations of all sets \( T'' \) of order \( |T''| \). Thus, given \( L \), accumulating the model outputs \( v^{\text{or}}_{N\setminus L} \) corresponding to all \( T\supseteq L \), then \( \sum\nolimits_{T: T \cap S \neq \emptyset, T \supseteq L} (-1)^{\vert T \vert - \vert L \vert} v^{\text{or}}_{N \setminus L} = v^{\text{or}}_{N \setminus L} \cdot \sum_{T' \subseteq N\setminus S \setminus L} \underbrace{\sum\nolimits_{\vert T'' \vert = 0}^{\vert S \vert} C_{\vert S \vert }^{\vert T''\vert } (-1)^{\vert T' \vert + \vert T'' \vert} }_{=0} = 0 \). 

Please see the complete derivation of the following formula.
\begin{equation}
\begin{aligned}
\sum\nolimits_{T:T \cap S \neq \emptyset} I^{\text{or}}_T
        &= \sum\nolimits_{T:T \cap S \neq \emptyset} \left[- \sum\nolimits_{L \subseteq T} (-1)^{\vert T \vert - \vert L \vert} v^{\text{or}}_{N \setminus L} \right]\\
        &= - \sum\nolimits_{L \subseteq N} \sum\nolimits_{T: T \cap S \neq \emptyset, T \supseteq L} (-1)^{\vert T \vert - \vert L \vert} v^{\text{or}}_{N \setminus L} \\
        &=  - \left[\sum_{\vert T' \vert = 1}^{\vert S \vert} C_{\vert S \vert}^{\vert T' \vert} (-1)^{\vert T' \vert} \right] \cdot \underbrace{u^{\text{or}}_S}_{L=N\setminus S} - \underbrace{v^{\text{or}}_\emptyset}_{L=N} \\
        &\quad- \sum_{L \cap S \neq \emptyset, L \neq N} \left[\sum_{T' \subseteq N\setminus S \setminus L} \left( \sum_{\vert T'' \vert = 0}^{\vert S \vert-\vert S \cap L \vert} C_{\vert S \vert - \vert S \cap L \vert}^{\vert T''\vert } (-1)^{\vert T' \vert + \vert T'' \vert} \right) \right]\cdot v^{\text{or}}_{N \setminus L}  \\
        &\quad- \sum_{L \cap S=\emptyset, L \neq N \setminus S} \left[ \sum_{T' \subseteq N\setminus S \setminus L} \left( \sum_{\vert T'' \vert=0}^{\vert S \vert} C_{\vert S \vert}^{\vert T'' \vert} (-1)^{\vert T' \vert + \vert T'' \vert}\right) \right] \cdot v^{\text{or}}_{N \setminus L}  \\
        &=  - (-1) \cdot u^{\text{or}}_S - v^{\text{or}}_\emptyset - \sum_{L \cap S \neq \emptyset, L \neq N} \left[\sum_{T' \subseteq N\setminus S \setminus L} 0 \right]\cdot v^{\text{or}}_{N \setminus L}  \\
        &\quad- \sum_{L \cap S=\emptyset, L \neq N \setminus S}\left[\sum_{T' \subseteq N\setminus S \setminus L} 0 \right] \cdot v^{\text{or}}_{N \setminus L}  \\
        &= u^{\text{or}}_S - v^{\text{or}}_\emptyset\\
        &= u^{\text{or}}_S
\end{aligned} 
\end{equation}

\textbf{(3) Universal matching theorem of AND-OR interactions.}
With the universal matching theorem of AND interactions and the universal matching theorem of OR interactions, we can easily get \( v(\boldsymbol{x}_S) = u^{\text{and}}_S + u^{\text{or}}_S 
= v(\boldsymbol{x}_\emptyset) + \sum_{\emptyset \neq T\subseteq S} I^{\text{and}}_T + \sum_{T: T\cap S \neq \emptyset} I^{\text{or}}_T \), thus, we obtain the universal matching theorem of AND-OR interactions.

\end{proof}

\subsection{Empirical Verification of Universal Matching on LLMs}
\label{sec:apdx-llm-universal-matching}

Given an input prompt $\boldsymbol{x}$, the above subsection theoretically shows that the LLM output function $v(\cdot)$ can be faithfully matched by the corresponding logical model function $\phi(\cdot)$ over all $2^n$ different masked states in $\Psi$. In this subsection, we further provide empirical evidence to verify this universal matching property on large language models.

\begin{figure*}[t]
    \centering
    \includegraphics[width=\textwidth]{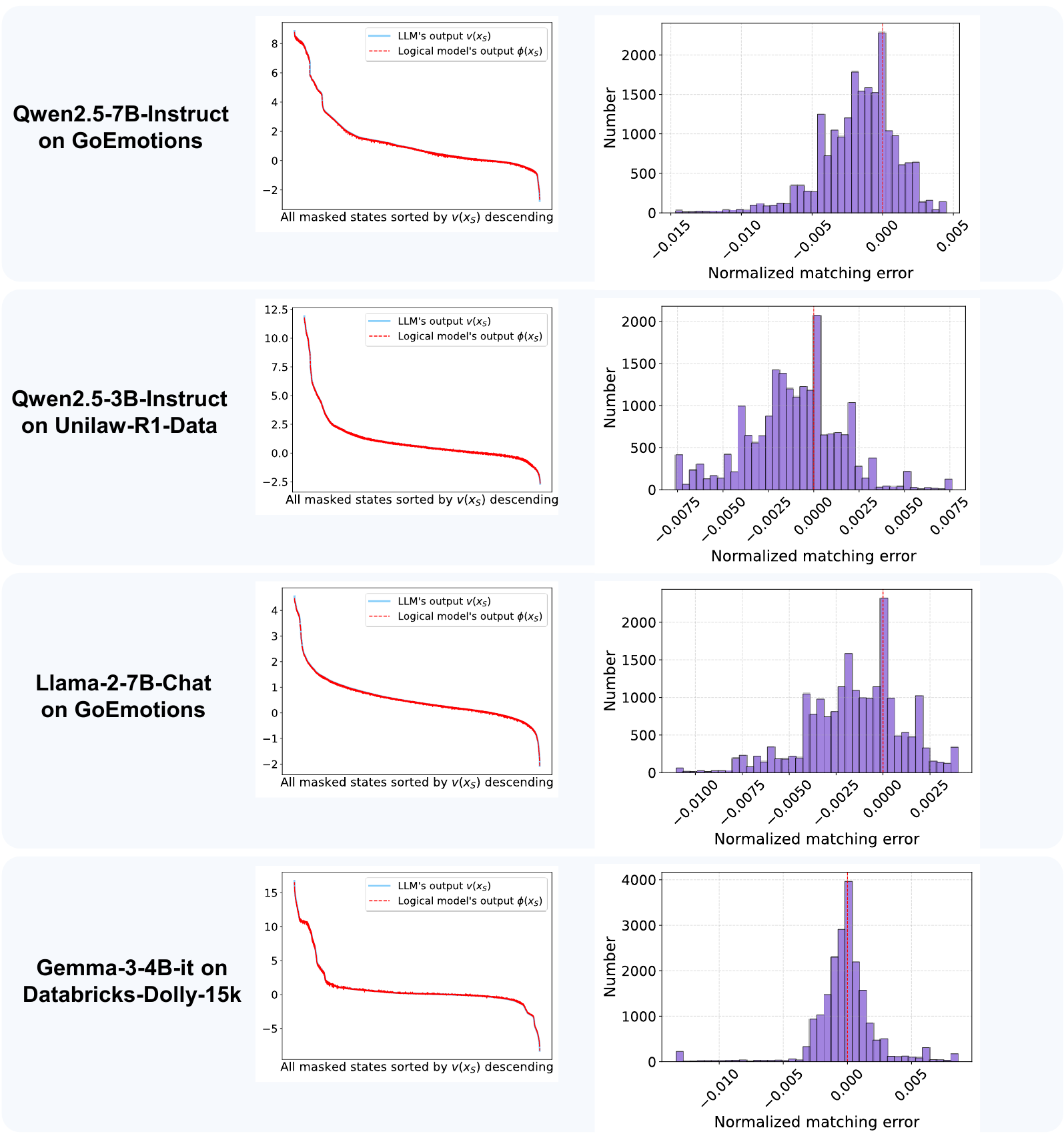}
    \caption{
    Empirical verification of universal matching on LLMs. 
    Each row corresponds to a different model-dataset setting. 
    In the left column, we aggregate the masked states from all selected samples and sort them in descending order according to the LLM output $v(\boldsymbol{x}_S)$. 
    The logical model output $\phi(\boldsymbol{x}_S)$ is plotted at the corresponding sorted positions, showing that the logical model closely matches the LLM output across different masked states. 
    In the right column, we report the histogram of the normalized matching error 
    $\epsilon(\boldsymbol{x}_S)=
    (\phi(\boldsymbol{x}_S)-v(\boldsymbol{x}_S))/
    |v(\boldsymbol{x})-v(\boldsymbol{x}_{\emptyset})|$ 
    over different masked states and input samples. 
    The errors are concentrated around zero across different model-dataset settings, further demonstrating the high fidelity of the logical model in approximating LLM outputs.
    }
    \label{img::appendix-universal}
\end{figure*}

Specifically, for each input sample $\boldsymbol{x}$, we compute the outputs of both the LLM and the corresponding logical model on all masked states $\boldsymbol{x}_S \in \Psi$. We then sort these masked states according to the LLM output $v(\boldsymbol{x}_S)$ in descending order, and compare the logical model output $\phi(\boldsymbol{x}_S)$ at the corresponding positions. As shown in Figure~\ref{img::appendix-universal}, the logical model closely matches the output curve of the LLM across different masked states, indicating that the logical model can accurately approximate the behavior of the LLM function.

In addition, we define the matching error as
\begin{equation}
\epsilon(\boldsymbol{x}_S)
=
\frac{
\phi(\boldsymbol{x}_S)-v(\boldsymbol{x}_S)
}{
\left|v(\boldsymbol{x})-v(\boldsymbol{x}_{\emptyset})\right|
},
\label{eq:appendix-llm-matching-error}
\end{equation}
where the denominator $\left|v(\boldsymbol{x})-v(\boldsymbol{x}_{\emptyset})\right|$ normalizes the error by the output scale of the LLM on the given sample. Figure~\ref{img::appendix-universal} shows the histogram of matching errors over different masked states and different input samples. The distribution is highly concentrated around zero, demonstrating that the matching error remains very small across samples and masking states. These empirical results further verify the high fidelity of the logical model in approximating LLM outputs.

We conduct experiments on multiple LLMs and datasets, including Llama2-7B-Chat and Qwen2.5-7B-Instruct on the GoEmotions dataset, Qwen2.5-3B-Instruct on the Unilaw-R1-Data dataset, and Gemma-3-4B-it on the Databricks-Dolly-15k dataset. For each model-dataset setting, we compute interaction-based logical models and compare their outputs with the corresponding LLM outputs over all masked states of the selected samples. The results consistently show that the logical models can faithfully reproduce the LLM output functions, providing strong empirical evidence for the universal matching ability of our interaction-based explanations.

\section{Validation of the Sparsity of Interactions}\label{Appendix::interaction-sparsity}
\citet{ren2024we} have proved three sufficient conditions for the sparsity of AND interactions. Accordingly, in this section, we first revisit the three conditions and then empirically validate the sparsity of interactions.

\subsection{Common Conditions for Sparse Interactions}
\label{sec:apdx-condition-for-sparsity}

\textbf{Condition 1.} \textit{The DNN does not encode extremely high-order interactions: {\small$\forall \ T\in \{T\subseteq N | \vert T\vert \ge M+1\}, \ I^{\text{\rm and}}_T =0$}.}

Condition 1 is common because extremely high-order interactions usually represent very complex and over-fitted patterns, which are unlikely to be learned by a well-trained DNN in real scenarios.

\textbf{Condition 2.} \textit{Let {\small$\bar{u}^{(k)}\overset{\text{\rm def}}{=}\mathbb{E}_{|S|=k}[v(\boldsymbol{x}_S)-v(\boldsymbol{x}_\emptyset)]$} denote the average classification confidence of the DNN over all masked samples $\boldsymbol{x}_S$ with $k$ unmasked input variables. This average classification confidence monotonically increases when $k$ increases: $\forall \ k' \le k$, {\small$\bar{u}^{(k')} \le \bar{u}^{(k)}$}.}

Condition 2 implies that a well-trained DNN is likely to have higher average classification confidence for less masked input samples.

\textbf{Condition 3.} \textit{Given the average classification confidence $\bar{u}^{(k)}$ of samples with $k$ unmasked input variables, there is a polynomial lower bound for the average classification confidence with $k' (k'\le k)$ unmasked input variables: {\small $\forall \ k' \le k, \ \bar{u}^{(k')} \ge (\frac{k'}{k})^p \ \bar{u}^{(k)}$}, where $p>0$ is a constant.}

Condition 3 suggests that the classification confidence of the DNN remains relatively stable even when presented with masked input samples. In real-world applications, the classification or detection of masked or occluded samples frequently occurs. As a result, a well-trained DNN typically develops the ability to classify such masked inputs by leveraging local information, which can be derived from the visible portions of the input. Consequently, the model should not produce a substantially reduced confidence score for masked samples.

\subsection{Empirical Verification of Interaction Sparsity on LLMs}
\label{sec:apdx-llm-sparsity}

Given an input sample, the preceding subsection presents sufficient conditions for interaction sparsity. In this subsection, we further provide empirical evidence showing that the interactions extracted from large language models are sparse in practice.

Specifically, we randomly select 20 input samples and extract the interactions encoded by the corresponding LLMs for these samples. Following \citep{ren2024we}, we normalize the interaction strengths by the maximum absolute interaction strength within each model-dataset setting, and sort all interactions in descending order according to their absolute values. We aggregate the interactions from all selected samples into a single pool and display them together in one figure for each model-dataset setting.

We conduct experiments on multiple LLMs and datasets, including Llama2-7B-Chat and Qwen2.5-7B-Instruct on the GoEmotions dataset, Qwen2.5-3B-Instruct on the Unilaw-R1-Data dataset, and Gemma-3-4B-it on the Databricks-Dolly-15k dataset. For each model-dataset setting, we compute the interaction strengths for the selected samples, merge all interactions across these samples, and analyze their sorted distributions.

\begin{figure*}[t]
    \centering
    \includegraphics[width=\textwidth]{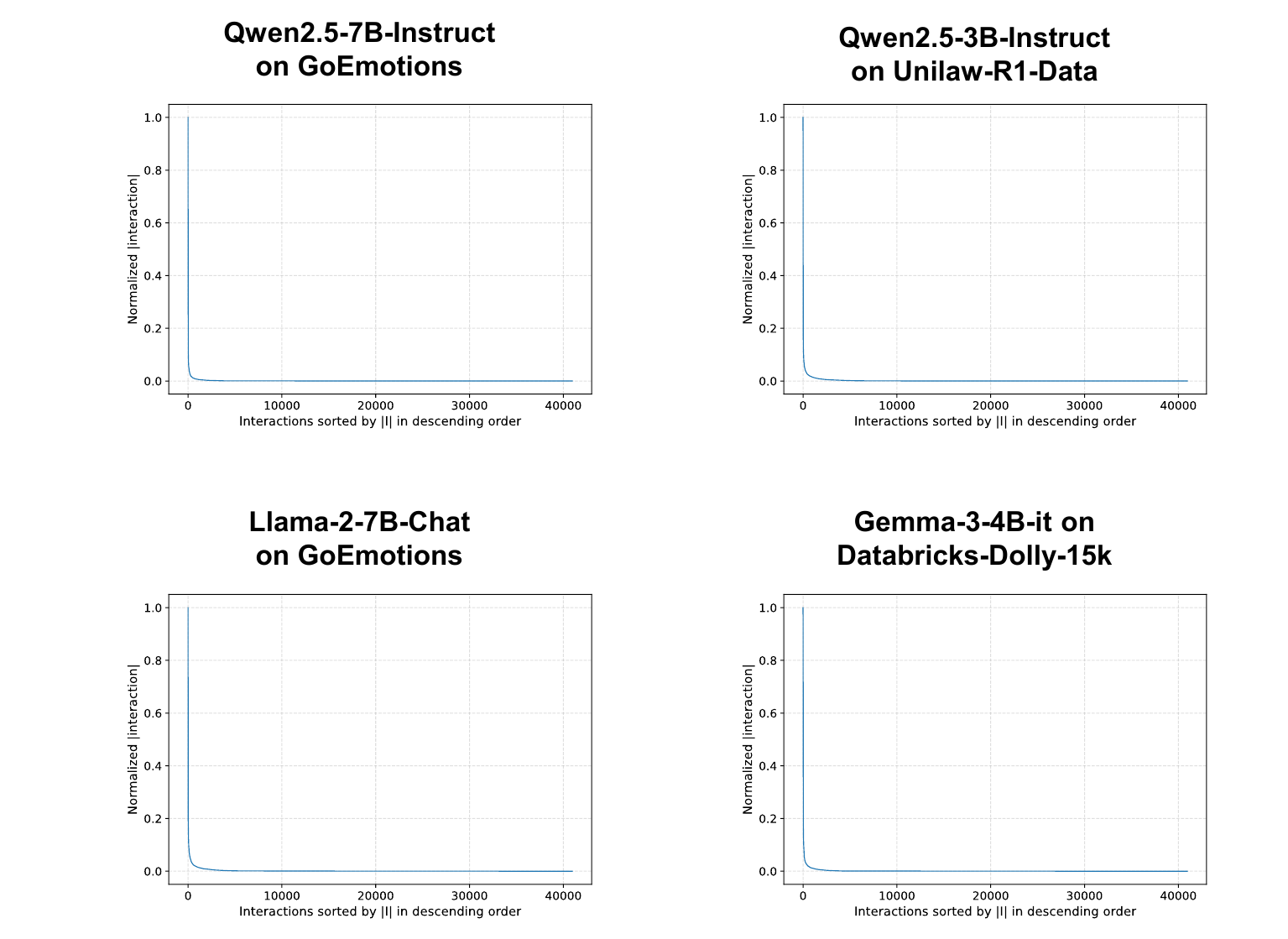}
    \caption{
    Empirical verification of interaction sparsity on LLMs. 
    We aggregate the interactions extracted from all selected samples into a single pool, normalize their strengths, and sort them in descending order according to their absolute values. 
    The results show that only a small fraction of interactions have relatively large magnitudes, while the vast majority are close to zero. 
    These results provide strong empirical evidence that the interactions are highly sparse in practice.
    }
    \label{img::appendix-sparsity}
\end{figure*}

As shown in Figure~\ref{img::appendix-sparsity}, only a small number of interactions have relatively large magnitudes, while the vast majority of interactions are close to zero. This phenomenon consistently holds across different LLMs and datasets, providing strong empirical evidence that the interactions are highly sparse in practice.

% \subsection{Empirical validation}
% Given an input sample, the preceding subsection presents sufficient conditions for interaction sparsity. Detailed proofs are provided in \citet{ren2024we}. In this subsection, we further present empirical evidence demonstrating that interactions are sparse in practice.

% Specifically, we randomly select a subset of input samples and extract the interactions encoded by the DNNs for these samples. Following \citet{ren2024we}, we normalize the interaction strengths and sort them in descending order. We conduct experiments by training VGG-11 and ResNet-20 on the CIFAR-10 dataset, AlexNet and VGG-16 on the Tiny-ImageNet dataset, and BERT-Base and BERT-Tiny on the SST-2 dataset. For each dataset, we randomly select five samples and analyze the normalized interaction distributions for each sample.

% As shown in Figures~\ref{img::appendix-sparsity}, the logical models encode only a small number of salient interactions for each input sample; that is, only a few interactions exhibit large effects, while the vast majority have effects close to zero. These results provide strong empirical evidence for the sparsity of interactions.

\section{Limitations}\label{Appendix::limitations}
Similar to the computation of Shapley values~\citep{shapley1953value}, interaction extraction incurs exponential computational cost in the worst case, as it requires evaluating masked states over the input. Nevertheless, prior work has shown that this cost can be substantially mitigated in practice by exploiting the sparsity of interactions and adopting efficient approximation algorithms.

For example, the Sparse M\"obius Transform reduces the complexity to $O(nK\log n)$ when only $K$ salient interactions exist~\citep{kang2024learning}. Spectral Explainer efficiently identifies important interactions for LLMs with long input sequences~\citep{kang2025spex}, while ProxySPEX further improves scalability by using lightweight proxy models to approximate the local decision behavior of the original model~\citep{butler2025proxyspex}. Together with practical engineering techniques~\citep{li2023defining}, these methods provide a practical pathway toward scalable interaction-based explanations.

In addition, for LLMs, interactions do not necessarily need to be extracted at the level of individual tokens. Depending on the analysis objective, one can compute interactions among phrases, words, or even sentences, which effectively reduces the number of input variables and substantially alleviates the sequence-length issue. Moreover, analyzing the evolution of interactions during the SFT process typically requires estimating interaction distributions rather than exhaustively enumerating all possible interactions. In such cases, selectively computing interactions over a small set of salient input variables is often sufficient to obtain a stable interaction distribution for generalization analysis. Consequently, the computational cost can be handled efficiently in typical experimental settings.

Therefore, although exact interaction extraction remains computationally expensive in theory, existing approximation algorithms, flexible choices of interaction granularity, and practical engineering techniques make interaction-based explanations increasingly feasible for modern large-scale models. Interaction-based explanations have also been applied to diverse DNNs, including LLMs, quantitative trading models, and pedestrian detection models in autonomous driving, as well as industrial model-evaluation scenarios~\citep{symtrustai2026}.

\section{More Experimental Results}\label{Appendix-more-results}
\subsection{Additional Results on the Evolution of Interactions}
\label{app:evolution-more-llms}
In this appendix, we provide additional results on the evolution of interactions during SFT across more LLMs and datasets. Figure~\ref{fig:additional-evolution-interactions} presents results for Llama-2-7B-Chat and Qwen2.5-3B-Instruct on GoEmotions, Qwen2.5-3B-Instruct and Qwen2.5-7B-Instruct on Unilaw-R1-Data, and Gemma-3-4B-it on Databricks-Dolly-15k. These results are consistent with the observations in the main paper. In the early denoising stage, LLMs rapidly remove many mutually canceling interactions while preserving only a small number of low-order interactions. In the later overfitting stage, many high-order mutually canceling interactions gradually emerge again.

\begin{figure}[t]
    \centering
    \includegraphics[width=0.99\textwidth]{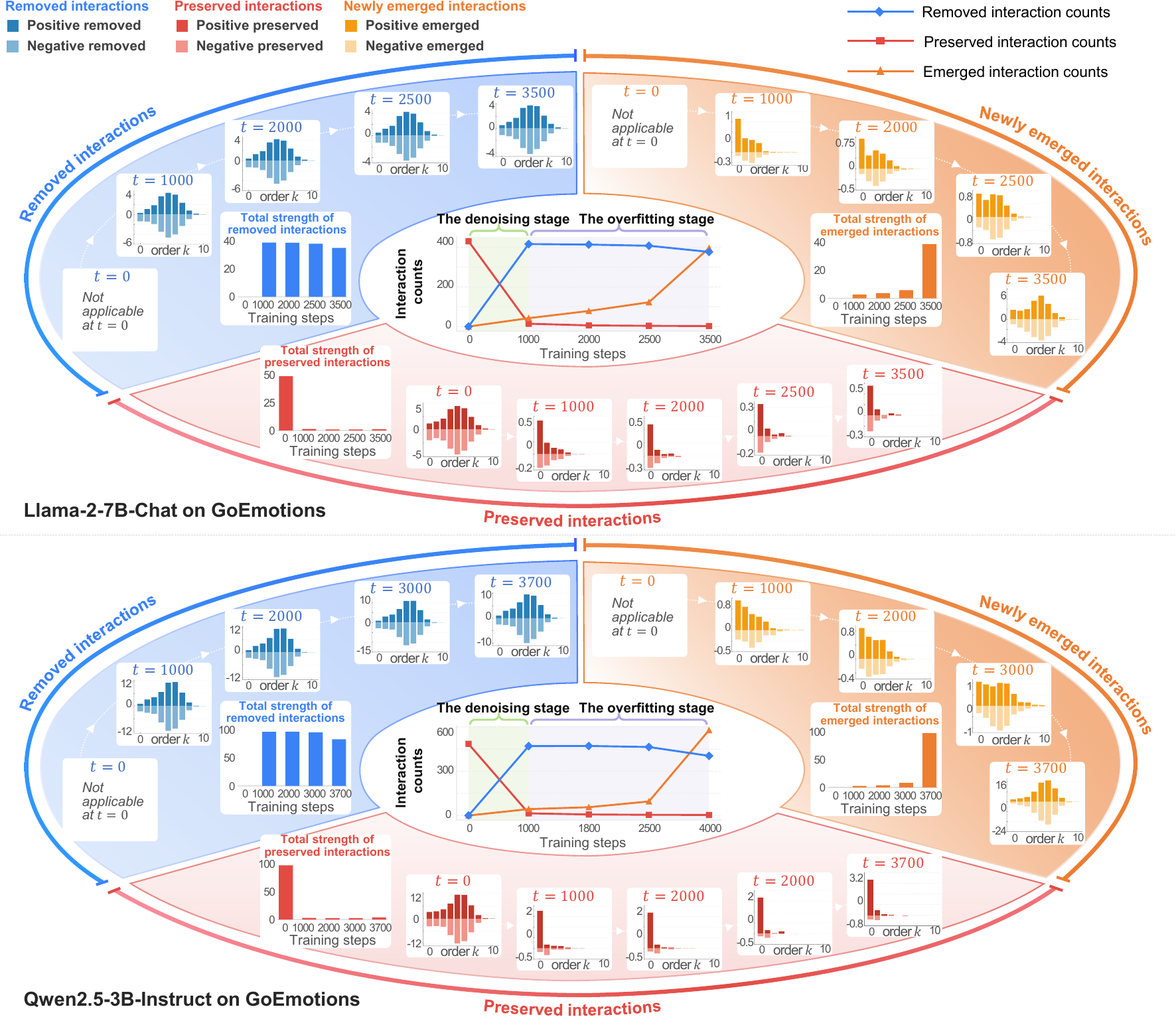}
    \caption{
    Additional results on the evolution of newly emerged, removed, and preserved interactions during SFT. 
    The figure shows Llama-2-7B-Chat and Qwen2.5-3B-Instruct on GoEmotions. 
    Additional results are presented on the next pages.
    }
    \label{fig:additional-evolution-interactions}
\end{figure}

\begin{figure}[t]
    \ContinuedFloat
    \centering
    \includegraphics[width=0.99\textwidth]{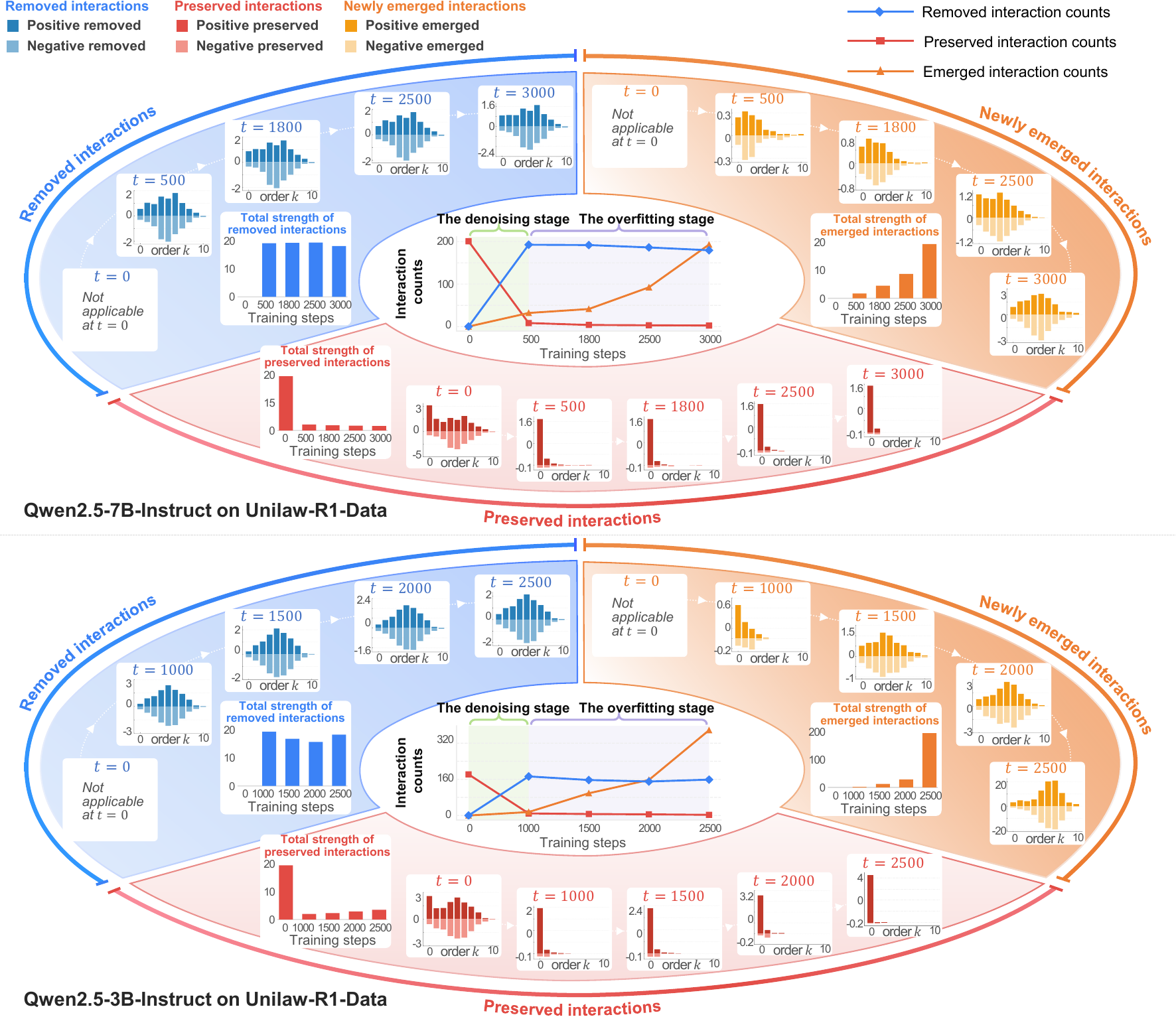}
    \caption{
    Additional results on the evolution of newly emerged, removed, and preserved interactions during SFT. 
    The figure shows Qwen2.5-3B-Instruct and Qwen2.5-7B-Instruct on Unilaw-R1-Data. 
    Additional results are presented on the next page.
    }
\end{figure}

\begin{figure}[t]
    \ContinuedFloat
    \centering
    \includegraphics[width=0.99\textwidth]{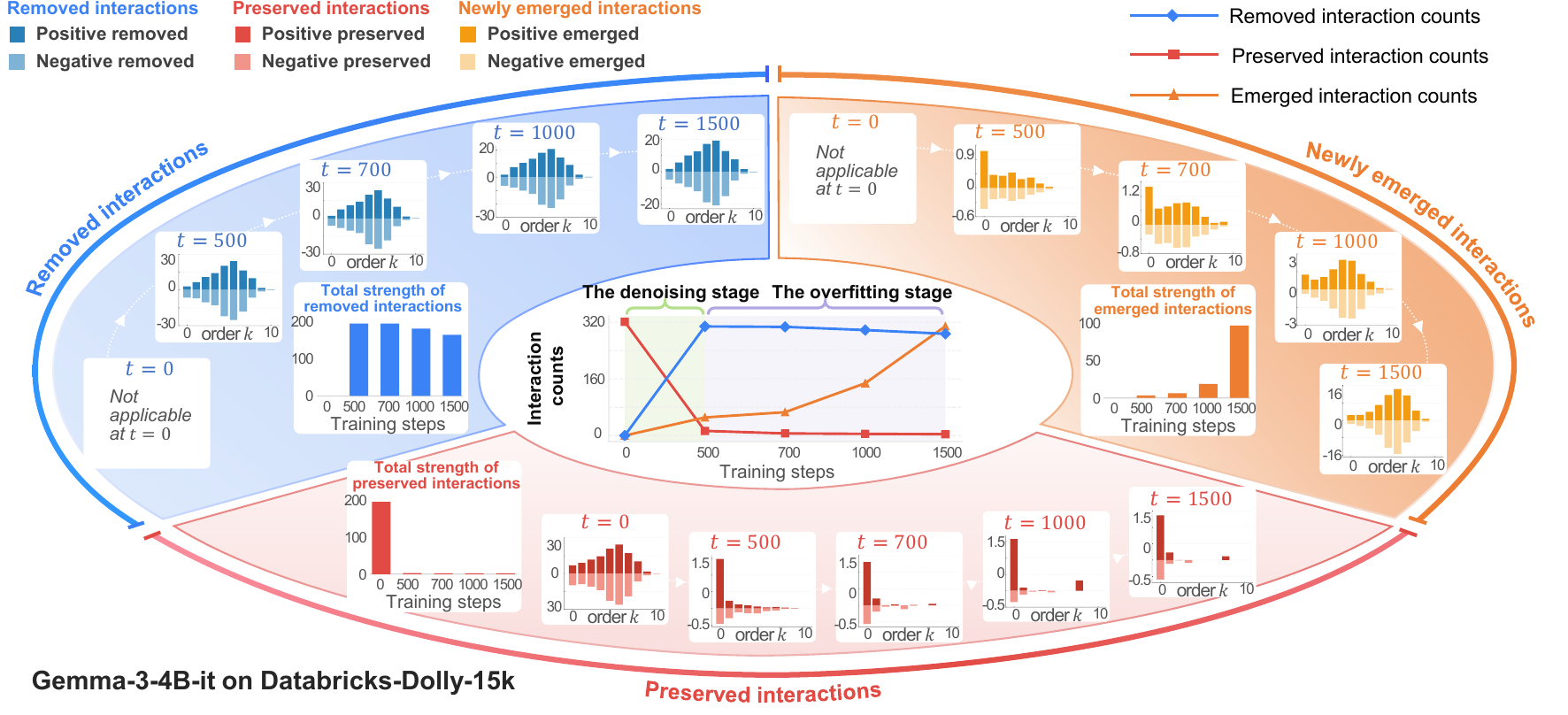}
    \caption{
    Additional results on the evolution of newly emerged, removed, and preserved interactions during SFT. 
    The figure shows Gemma-3-4B-it on Databricks-Dolly-15k. 
    Across these LLMs and datasets, the results consistently show that SFT first removes many mutually canceling interactions in a short denoising stage, while many high-order mutually canceling interactions emerge again in the later overfitting stage.
    }
\end{figure}
% \begin{figure}[p]
%     \centering
%     \includegraphics[width=0.99\textwidth]{figs/Appendix-distribution1.pdf}
%     \caption{
%     Additional results on the evolution of newly emerged, removed, and preserved interactions during SFT. 
%     The figure shows Llama-2-7B-Chat and Qwen2.5-3B-Instruct on GoEmotions. 
%     Additional results are presented on the next page.
%     }
%         \label{fig:additional-evolution-interactions}
% \end{figure}

% \begin{figure}[p]
%     \centering
%     \includegraphics[width=0.99\textwidth]{figs/Appendix-distribution2.pdf}
%     \caption{
%     Additional results on the evolution of newly emerged, removed, and preserved interactions during SFT. 
%     The figure shows Qwen2.5-3B-Instruct and Qwen2.5-7B-Instruct on Unilaw-R1-Data. 
%     Additional results are presented on the next page.
%     }
% \end{figure}

% \begin{figure}[p]
%     \ContinuedFloat
%     \centering
%     \includegraphics[width=0.99\textwidth]{figs/Appendix-distribution3.pdf}
%     \caption{
%     Additional results on the evolution of newly emerged, removed, and preserved interactions during SFT. 
%     This figure shows Gemma-3-4B-it on Databricks-Dolly-15k. 
%     Across these LLMs and datasets, the results consistently show that SFT first removes many mutually canceling interactions in a short denoising stage, while many high-order mutually canceling interactions emerge again in the later overfitting stage.
%     }
% \end{figure}

\subsection{Additional Results on the Representation Quality of Interactions}
\label{app:quality-more-llms}
\begin{figure}[!t]
    \centering
    \includegraphics[width=0.99\textwidth]{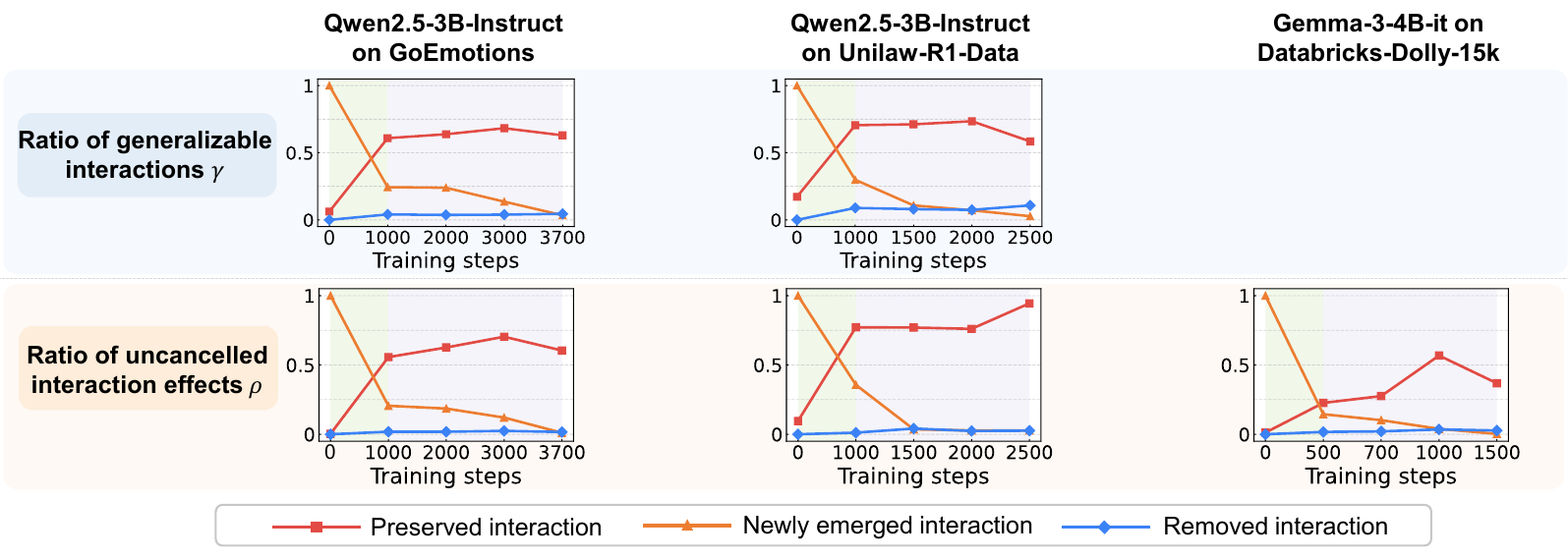}
    \caption{
    Additional results on the representation quality of newly emerged, removed, and preserved interactions during SFT. 
    The figure shows Qwen2.5-3B-Instruct on GoEmotions, Qwen2.5-3B-Instruct on Unilaw-R1-Data, and Gemma-3-4B-it on Databricks-Dolly-15k. 
    Since no baseline LLM is trained on Databricks-Dolly-15k, interaction generalizability $\gamma$ is not evaluated for Gemma-3-4B-it, and only the uncancelled-effect ratio $\rho$ is reported.
    }
    \label{fig:additional-quality-interactions}
\end{figure}

In this appendix, we provide additional results on the representation quality of newly emerged, removed, and preserved interactions during SFT. Figure~\ref{fig:additional-quality-interactions} reports results for Qwen2.5-3B-Instruct on GoEmotions, Qwen2.5-3B-Instruct on Unilaw-R1-Data, and Gemma-3-4B-it on Databricks-Dolly-15k. These results are consistent with the observations in the main paper. Newly emerged interactions exhibit high representation quality only in the short denoising stage, but become less generalizable and more mutually canceling in the later stage. Removed interactions remain low-quality throughout SFT, while preserved interactions improve after noisy interactions are removed. Since only one LLM is fine-tuned on Databricks-Dolly-15k, no baseline LLM is available for measuring interaction generalizability $\gamma$ on this dataset. Therefore, we only report the uncancelled-effect ratio $\rho$ for Gemma-3-4B-it.

\subsection{Examples of AND-OR Logical Models Explaining LLMs}\label{Appendix::examples-LLM}
In this section, to demonstrate that AND-OR logical models can faithfully explain large language models, we present examples showing that, given specific input prompts, AND-OR logical models can be constructed to faithfully explain both the DeepSeek-r1-distill-llama-8B~\citep{guo2025deepseek} and Qwen2.5-7B~\citep{bai2023qwen} models, hereafter referred to as the DeepSeek and Qwen models, respectively. Figure \ref{fig:andor_models} illustrates the AND-OR logical models used to explain DeepSeek and Qwen under different input prompts.

\begin{figure}[t]
    \centering
    % sample0
    \includegraphics[width=0.85\textwidth]{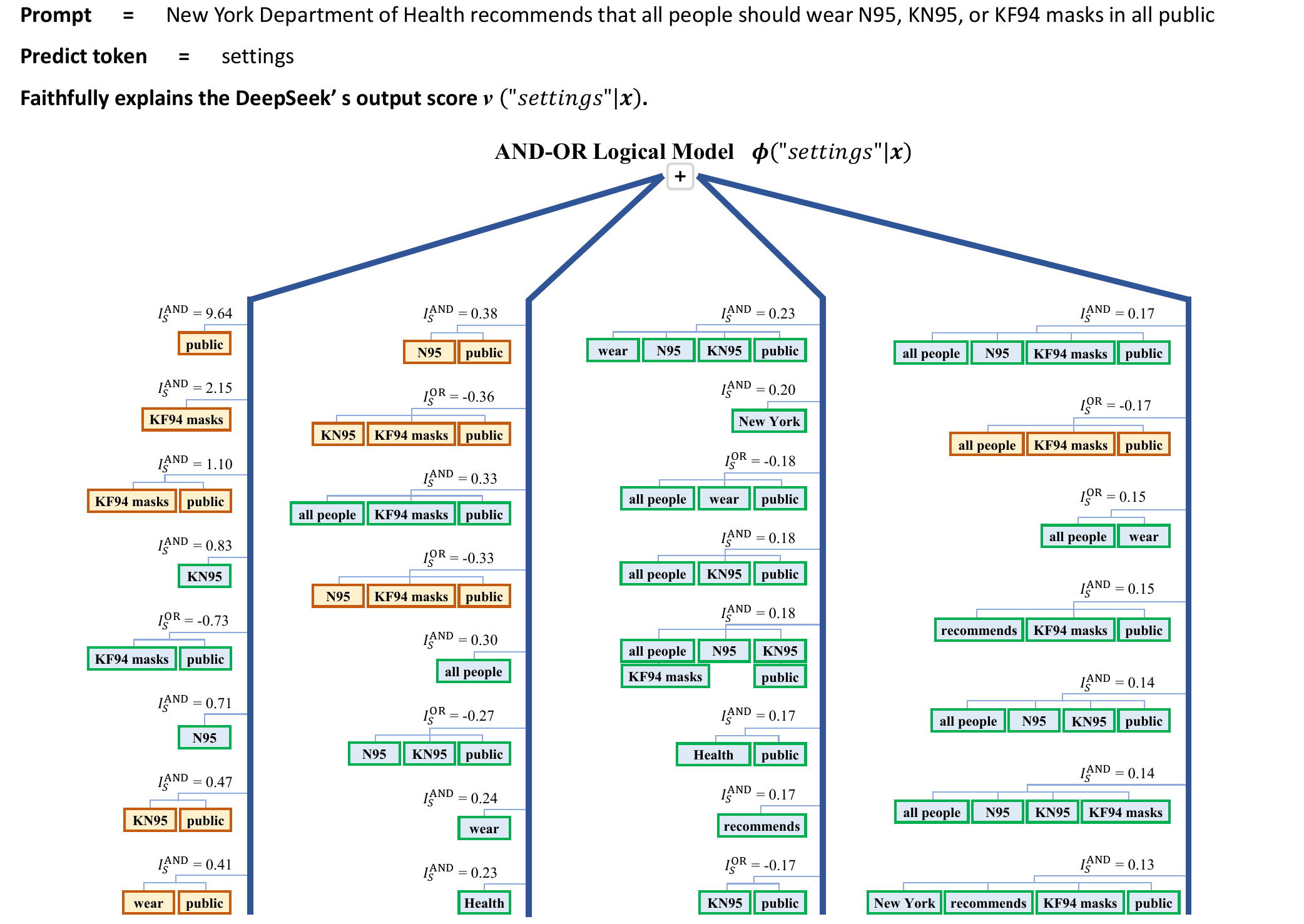}\vspace{30pt}
    \includegraphics[width=0.85\textwidth]{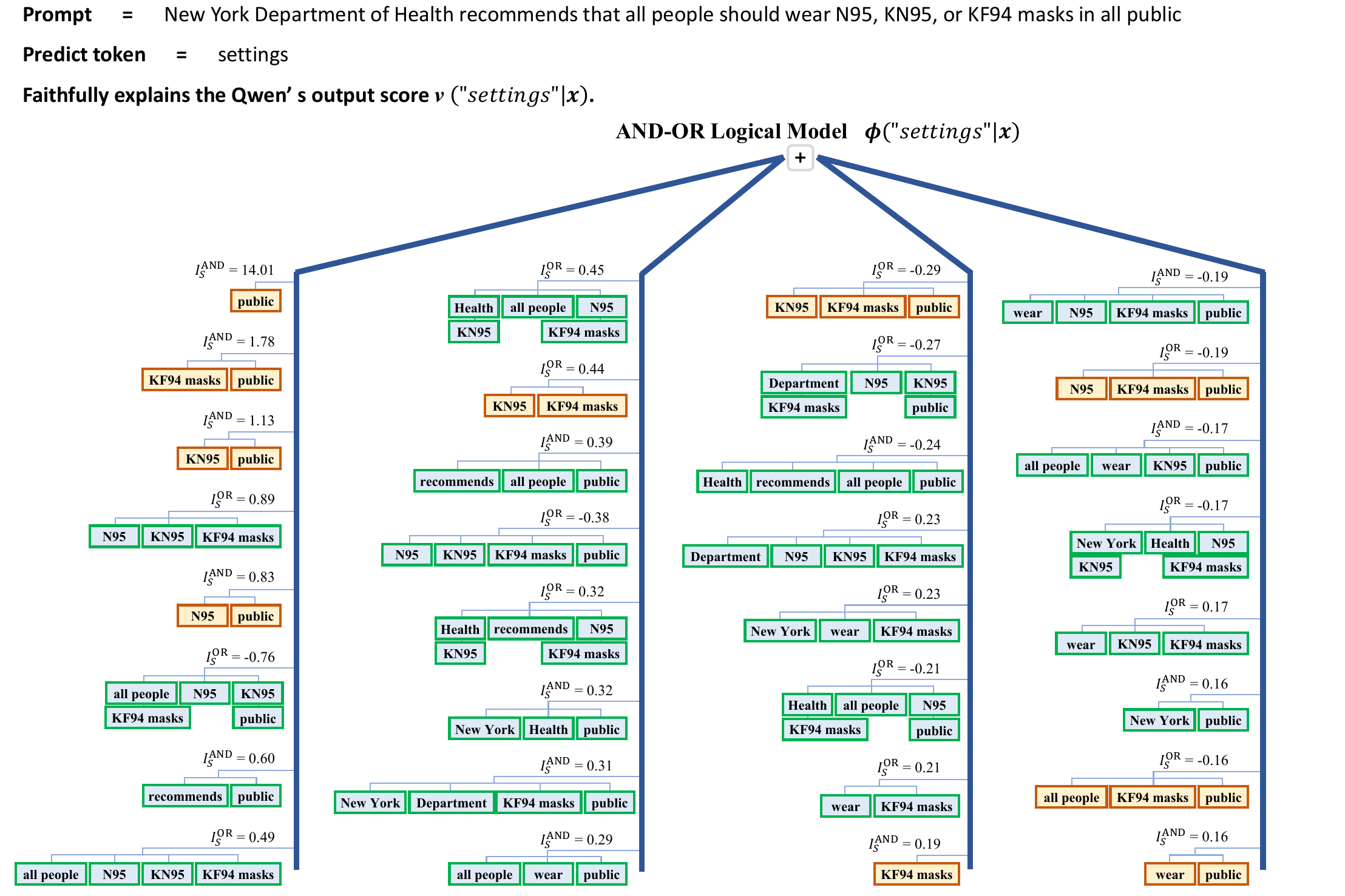}
    \caption{An example of AND-OR logical models constructed to faithfully explain the output scores of the DeepSeek model (top) and the Qwen model (bottom) on a single sample. A further example is presented on the next page.}
    \label{fig:andor_models}
\end{figure}

\begin{figure}[t]
    \ContinuedFloat
    \centering
    % sample1
    \includegraphics[width=0.85\textwidth]{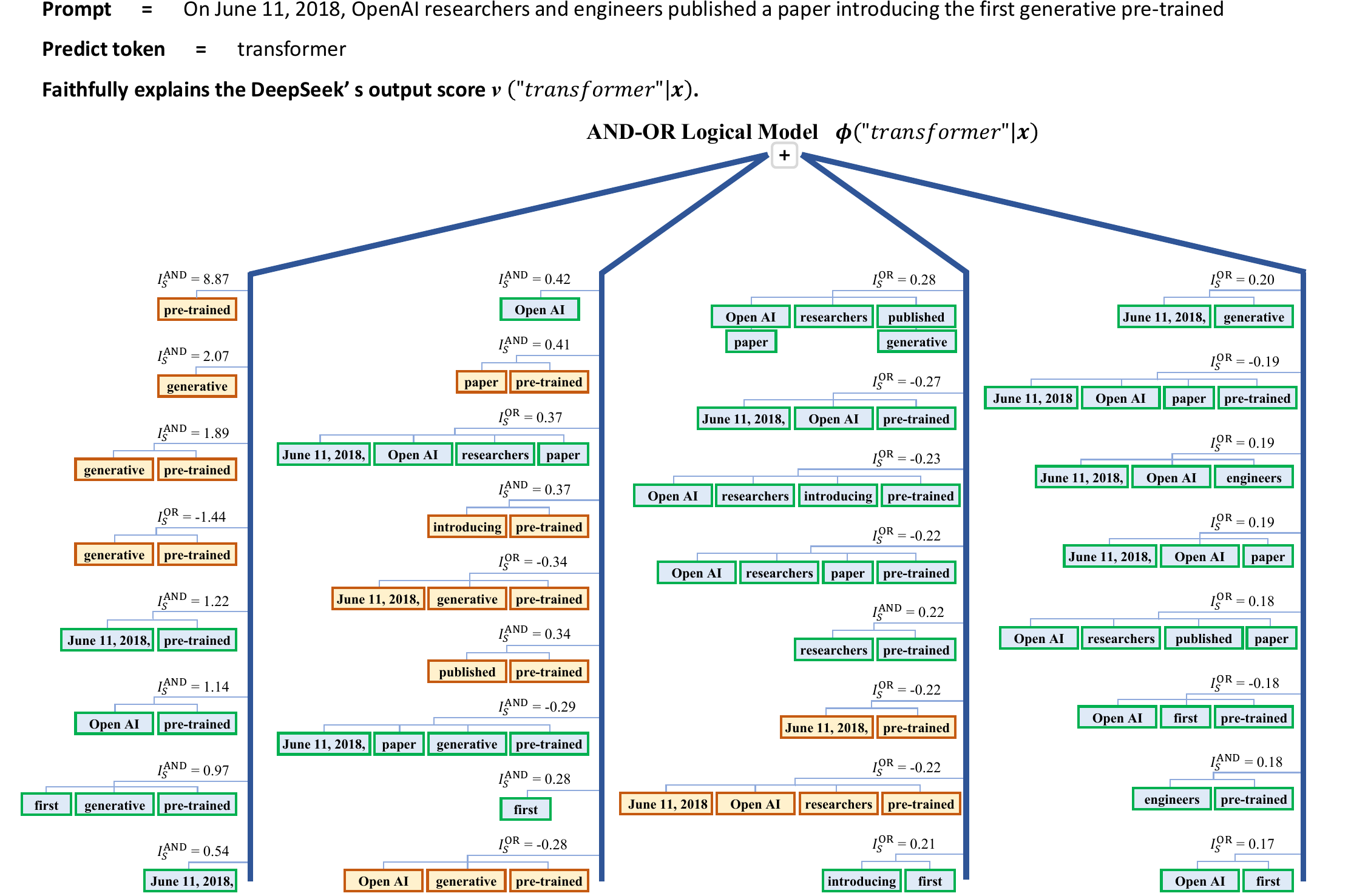}\vspace{30pt}
    \includegraphics[width=0.85\textwidth]{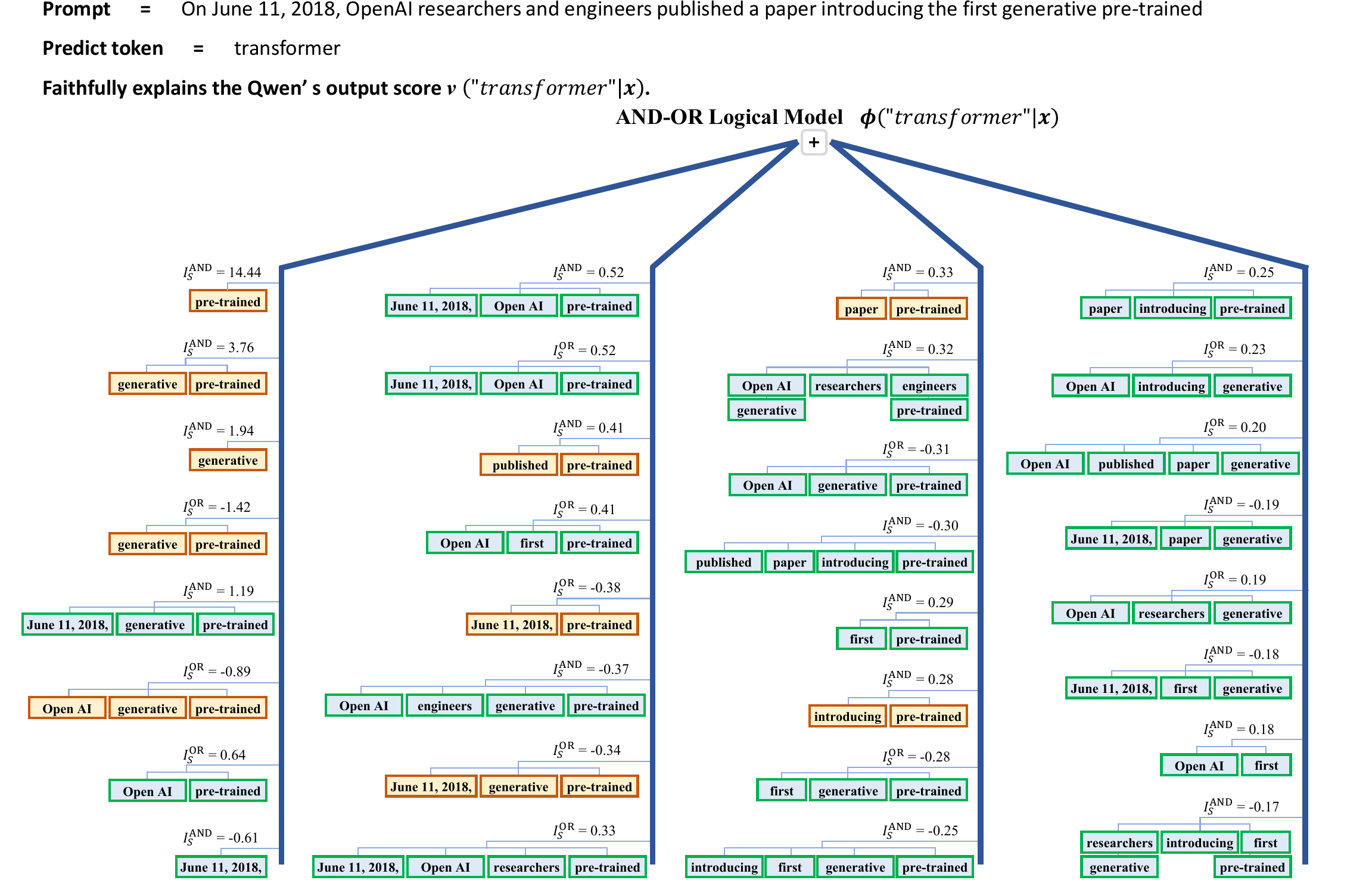}
    \caption{Another example of AND-OR logical models explaining the DeepSeek model (top) and the Qwen model (bottom) on a different sample.}
\end{figure}

\section{Experimental Details}\label{sec:experimental_setting}
\subsection{Training Settings}\label{sec:training_setting}
\textbf{Models and datasets.} We conduct supervised fine-tuning (SFT) experiments on multiple LLMs across different datasets. For the GoEmotions dataset~\citep{demszky2020goemotions}, we fine-tune Qwen2.5-3B-Instruct~\citep{qwen2.5}, Qwen2.5-7B-Instruct~\citep{qwen2.5}, Llama-2-7B-Chat~\citep{touvron2023llama}, and Llama-3-8B-Instruct~\citep{grattafiori2024llama}. For the Unilaw-R1-Data dataset~\citep{cai2025unilaw}, we fine-tune Qwen2.5-3B-Instruct and Qwen2.5-7B-Instruct. For the Databricks-Dolly-15k dataset~\citep{DatabricksBlog2023DollyV2}, we fine-tune Gemma-3-4B-it~\citep{gemma_2025}. All datasets are formatted into instruction-response style examples, and the corresponding model-specific chat template is used for each model.

\textit{GoEmotions.} 
GoEmotions~\citep{demszky2020goemotions} is a large-scale English emotion classification dataset collected from Reddit comments. It contains approximately 58K carefully curated comments annotated with 27 fine-grained emotion categories and a neutral label. Each instance consists of a text comment and one or more emotion labels, making the dataset suitable for both multi-class and multi-label emotion classification. The dataset is publicly available, and its hosting GitHub repository is released under the Apache License 2.0.

\textit{Unilaw-R1-Data.}
Unilaw-R1-Data~\citep{cai2025unilaw} is a legal-domain instruction dataset constructed for training legal reasoning models. The dataset is derived from objective question-answering entries in the legal domain, including JEC-QA and the authors' primary legal reasoning data, and is further distilled and partitioned into supervised fine-tuning and reinforcement learning subsets. The dataset is publicly available, but the original source does not specify an explicit dataset license.

\textit{Databricks-Dolly-15k.}
Databricks-Dolly-15k~\citep{DatabricksBlog2023DollyV2} is an open-source English instruction-following dataset consisting of more than 15K human-generated instruction-response records. The data were created by Databricks employees and cover multiple instruction categories, including brainstorming, classification, closed-form question answering, generation, information extraction, open-domain question answering, and summarization. The dataset is publicly available under the CC BY-SA 3.0 license.

\textbf{Fine-tuning details.} All SFT experiments are implemented with LLaMA-Factory~\citep{zheng2024llamafactory}. We adopt LoRA~\citep{hu2022lora} as the parameter-efficient fine-tuning method. Unless otherwise specified, the LoRA rank is set to $r=8$, and LoRA adapters are applied to all eligible linear modules. The maximum sequence length is set to 8192, and the maximum number of training samples is capped at 100,000 when applicable. During training, the loss is computed only on the labeled response tokens, while prompt tokens are excluded from the loss.

We use a peak learning rate of $1\times 10^{-4}$ with a cosine learning-rate scheduler and a warmup ratio of 0.1. Models are trained with bfloat16 precision for 30 epochs. The per-device training batch size is set to 1, and gradients are accumulated over 8 steps. All experiments are conducted on 8 NVIDIA Tesla V100-PCIE-32GB GPUs, resulting in an effective batch size of 64 samples per optimization step.

\subsection{Baseline LLMs for Measuring Interaction Generalizability.}
To quantify the generalizability of interactions, we introduce a baseline LLM $v'$ for each target LLM $v$, where $v'$ is fine-tuned on the same dataset but has a different architecture from $v$. For the GoEmotions dataset, we use the fine-tuned Llama-2-7B-Chat model as the baseline LLM for the Qwen2.5-3B-Instruct, Qwen2.5-7B-Instruct, and Llama-3-8B-Instruct target models. When the target model is Llama-2-7B-Chat, we instead use the fine-tuned Qwen2.5-3B-Instruct model as the baseline LLM. For the Unilaw-R1-Data dataset, since we fine-tune two LLMs, Qwen2.5-3B-Instruct and Qwen2.5-7B-Instruct, we use them as the baseline LLMs for each other. For the Databricks-Dolly-15k dataset, we fine-tune only one LLM, Gemma-3-4B-it, and therefore do not evaluate the generalizability of interactions on this dataset.

When plotting the trajectories of $\gamma$ and $\rho$, 
newly emerged interactions and removed interactions are not available before SFT because they are defined by their changes along the SFT trajectory. For consistency of visualization, we use conventional boundary values for such undefined entries, setting $\gamma=\rho=1$ for newly emerged interactions and $\gamma=\rho=0$ for removed interactions. These boundary values only serve to anchor the curves and do not affect our quantitative comparisons.

\subsection{Computer Resources}\label{sec:resources}
All LLMs can be trained on 8 NVIDIA Tesla V100-PCIE-32GB GPUs, and can be finished within 120 hours. Calculating all interactions for an input sample takes an average of 30-50 seconds.

\subsection{Details About How to Calculate Interactions for Different LLMs}\label{sec:players}
Interaction-based explanations have been widely applied to real-world DNNs, largely due to prior efforts~\citep{ren2024towards, li2023does} that substantially reduce the computational cost of interaction extraction. Moreover, analyses of the evolution of interactions during the SFT of LLMs typically require computing only interaction distributions rather than exhaustively enumerating all interactions. As a result, selectively computing interactions over a small number of salient input variables is usually sufficient to obtain stable interaction distributions for analysis. Consequently, the computation can be handled efficiently in typical experimental settings.

Extensive experiments on large language models further support this observation. We find that LLMs usually attend to only a small number of salient text segments  when making predictions. As a result, explanations based on fewer than 10 salient text segments are often sufficient to faithfully capture the LLM’s inference patterns. This empirical property significantly improves the computational efficiency of interaction-based explanations and ensures their practical applicability to large-scale models and datasets.

Specifically, we followed~\citep{chen2024defining} to define each input variable as the embedding vector(s) of a word and calculate interactions. Besides, the distribution of interactions is computed by averaging the distribution across different samples. Furthermore, we followed~\citep{chen2024defining} to use the first target token of the response for interaction analysis.

\begin{algorithm}
\caption{Compute AND and OR Interactions and Select Salient Ones}
\label{alg:compute-interactions}
\begin{algorithmic}[1]
\REQUIRE Deep neural network $v$, input sample $\boldsymbol{x} = [x_1, x_2, \ldots, x_n]^T$, set of indices $N = \{1, 2, \ldots, n\}$, small noise threshold $\zeta$, significance threshold $\tau$, convergence threshold $\epsilon$.
\ENSURE AND interactions $I^{\text{and}}_T$, OR interactions $I^{\text{or}}_T$, and interaction sets $\Omega^{\text{and}}$ and $\Omega^{\text{or}}$.
\STATE Initialize learnable parameters $\{\gamma_L\}$ and $\{\delta_L\}$ for all $L \subseteq N$.
\STATE Compute baseline output $v(\boldsymbol{x}_\emptyset)$, where $\boldsymbol{x}_\emptyset$ is the masked sample with all variables removed.
\STATE Initialize previous loss $\mathcal{L}_{\text{prev}} \gets \infty$.
\REPEAT
    \FOR{each subset $L \subseteq N$}
        \STATE Compute masked sample $\boldsymbol{x}_L$ by removing variables not in $L$.
        \STATE Compute network output $v(\boldsymbol{x}_L)$.
        \STATE Compute noise term $\delta_L$ constrained in $[-\zeta, \zeta]$, where $\zeta = 0.01 \cdot |v(\boldsymbol{x}) - v(\boldsymbol{x}_\emptyset)|$.
        \STATE Decompose $v(\boldsymbol{x}_L)$ into AND and OR components:
        \STATE $u^{\text{and}}_L \gets 0.5 \cdot (v(\boldsymbol{x}_L) - \delta_L) + \gamma_L$
        \STATE $u^{\text{or}}_L \gets 0.5 \cdot (v(\boldsymbol{x}_L) - \delta_L) - \gamma_L$
    \ENDFOR
    \FOR{each subset $T \subseteq N$}
        \STATE Compute AND interaction $I^{\text{and}}_T$:
        $$
        I^{\text{and}}_T \gets \sum_{L \subseteq T} (-1)^{|T| - |L|} u^{\text{and}}_L
        $$
        \STATE Compute OR interaction $I^{\text{or}}_T$:
        $$
        I^{\text{or}}_T \gets -\sum_{L \subseteq T} (-1)^{|T| - |L|} u^{\text{or}}_{N \setminus L}
        $$
    \ENDFOR
    \STATE Compute current loss $\mathcal{L} \gets \sum_{T \subseteq N} \left( |I^{\text{and}}_T| + |I^{\text{or}}_T| \right)$.
    \STATE Optimize parameters $\{\gamma_L\}$ and $\{\delta_L\}$ to minimize $\mathcal{L}$.
    \STATE Check for convergence: $|\mathcal{L} - \mathcal{L}_{\text{prev}}| < \epsilon$.
    \STATE Update previous loss: $\mathcal{L}_{\text{prev}} \gets \mathcal{L}$.
\UNTIL{convergence}
\STATE Select AND interactions:
$$
\Omega^{\text{and}} \gets \{T \subseteq N: |I^{\text{and}}_T| > \tau\}
$$
\STATE Select OR interactions:
$$
\Omega^{\text{or}} \gets \{T \subseteq N: |I^{\text{or}}_T| > \tau\}
$$

\STATE \textbf{return} $I^{\text{and}}_T$, $I^{\text{or}}_T$, $\Omega^{\text{and}}$, and $\Omega^{\text{or}}$.
\end{algorithmic}
\end{algorithm}

\section{Details to Extract the Sparsest AND-OR Interactions}
\label{sec:apdx-optimize-pq}

A method is proposed~\citep{li2023defining, chen2024defining} to simultaneously extract AND interactions $I^{\text{and}}_T$ and OR interactions $I^{\text{or}}_T$ from the network output. Given a masked sample $\boldsymbol{x}_L$, \citet{li2023defining} proposed to learn a decomposition $v(\boldsymbol{x}_L)=u^{\text{and}}_L + u^{\text{or}}_L$ towards the sparsest interactions. 
The component {$u^{\text{and}}_L$} was explained by AND interactions, and the component {$u^{\text{or}}_L$} was explained by OR interactions.
Specifically, they decomposed $v(\boldsymbol{x}_L)$ into $u^{\text{and}}_L= 0.5 \cdot v(\boldsymbol{x}_L)+\gamma_L$ and $u^{\text{or}}_L= 0.5 \cdot v(\boldsymbol{x}_L) -\gamma_L$, where $\{\gamma_L:L\subseteq N\}$ is a set of learnable variables that determine the decomposition. In this way, the AND interactions and OR interactions can be computed, \textit{i.e.}, $I^{\text{and}}_T=\sum\nolimits_{L \subseteq T}(-1)^{|T|-|L|} u^{\text{and}}_L$, and  $I^{\text{or}}_T=-\sum\nolimits_{L \subseteq T}(-1)^{|T|-|L|} v^{\text{or}}_{N \setminus L}$.

The parameters $\{\gamma_L\}$ were learned by minimizing the following LASSO-like loss to obtain sparse interactions:
\begin{equation}
\label{eq:loss-pq}
    \min_{\{\gamma_L\}} \sum_{T\subseteq N} \vert I^{\text{and}}_T \vert + \vert I^{\text{or}}_T \vert
\end{equation}

\textbf{Removing small noises.} A small noise $\delta$ in the network output may significantly affect the extracted interactions, especially for high-order interactions. Thus, ~\citep{li2023defining} proposed to learn to remove a small noise term $\delta_T$ from the computation of AND-OR interactions.
Specifically, the decomposition was rewritten as {$u^{\text{and}}_L=0.5 (v(\boldsymbol{x}_L) -\delta_L) +\gamma_L$} and {$u^{\text{or}}_L=0.5 (v(\boldsymbol{x}_L)-\delta_L) -\gamma_L$}.
Thus, the parameters {$\{\delta_L\}$} and {$\{\gamma_L\}$} are simultaneously learned by minimizing the loss function in Eq.~(\ref{eq:loss-pq}).
The values of {$\{\delta_L\}$} were constrained in $[-\zeta, \zeta]$ where {$\zeta=0.01\cdot \vert v(\boldsymbol{x})-v(\boldsymbol{x}_\emptyset) \vert$}. Therefore, we can also set  $\epsilon = \zeta$ to ensure the fidelity of using the logical model function $\phi(\cdot)$ to mimic the DNN function $v(\cdot)$.

\textbf{Algorithm of extracting AND-OR interactions.} The technical details of computing \(I_T^{\text{and}}\) and \(I_T^{\text{or}}\) are provided in the following pseudocode in Algorithm~\ref{alg:compute-interactions}.

\section{Broader Impacts}\label{sec:impacts}
This paper uses interactions to understand the inconsistent effects of SFT on LLMs. We find that: (1) SFT primarily removes noise-like interactions from LLMs, while rarely inducing reliable new interactions; and (2) this denoising stage is extremely brief, after which continued fine-tuning leads to the emergence of numerous overfitted patterns. Our findings provide a principled signal for early stopping and for monitoring representation quality during fine-tuning. This may help reduce unnecessary computation, lower the cost of model adaptation, and mitigate the risk of learning unreliable inference patterns. More broadly, our results challenge the assumption that longer fine-tuning or larger SFT datasets are always beneficial. 

% %%%%%%%%%%%%%%%%%%%%%%%%%%%%%%%%%%%%%%%%%%%%%%%%%%%%%%%%%%%%

% \newpage
% \input{checklist.tex}

\end{document}